\documentclass{article}
 
\PassOptionsToPackage{square, numbers}{natbib}

\usepackage[preprint]{neurips_2025}

\usepackage[utf8]{inputenc} 
\usepackage[T1]{fontenc}    
\usepackage{hyperref}       
\usepackage{url}            
\usepackage{booktabs}       
\usepackage{amsfonts}       
\usepackage{nicefrac}       
\usepackage{microtype}      
\usepackage{xcolor}         

\usepackage{graphicx}
\usepackage{caption}
\usepackage{times}
\usepackage{epsfig}

\usepackage{amsmath}
\usepackage{amssymb}
\usepackage{makecell}  

\usepackage{mathtools}
\usepackage{amsmath,bm}
\usepackage[noend]{algpseudocode}
\usepackage{algorithmicx,algorithm}

\usepackage{booktabs}
\usepackage{multirow}
\usepackage{adjustbox}
\usepackage{colortbl}
\definecolor{Gray}{gray}{0.9}
\usepackage{xcolor}
\usepackage{color,colortbl}

\usepackage{wrapfig}

\newcommand{\fref}[1]{Fig.~\ref{#1}}
\newcommand{\tref}[1]{Tab.~\ref{#1}}
\newcommand{\eref}[1]{Eq.~\ref{#1}}
\newcommand{\sref}[1]{Sec.~\ref{#1}}
\newcommand{\R}{\mathbb{R}}
\newcommand{\diff}{\mathop{}\!\mathrm{d}}

\title{Elucidating and Endowing the Diffusion Training Paradigm
 for General Image Restoration}
\author{%
  Xin Lu,~~~Xueyang Fu$^{\dagger}$,~~~Jie Xiao,~~~Zihao Fan,~~~Yurui Zhu,~~~Zheng-Jun Zha\\
  \normalsize
MoE Key Laboratory of Brain-inspired Intelligent Perception and Cognition,\\
  School of Information Science and Technology,   University of Science and Technology of China\\
\texttt{luxion@mail.ustc.edu.cn, xyfu@ustc.edu.cn} \\
}

\begin{document}

\maketitle

\vspace{-12mm}
\begin{figure}[ht]
\setlength{\abovecaptionskip}{0.2cm}
\setlength{\belowcaptionskip}{0.2cm}
  \centering
   \includegraphics[width=0.96\linewidth]{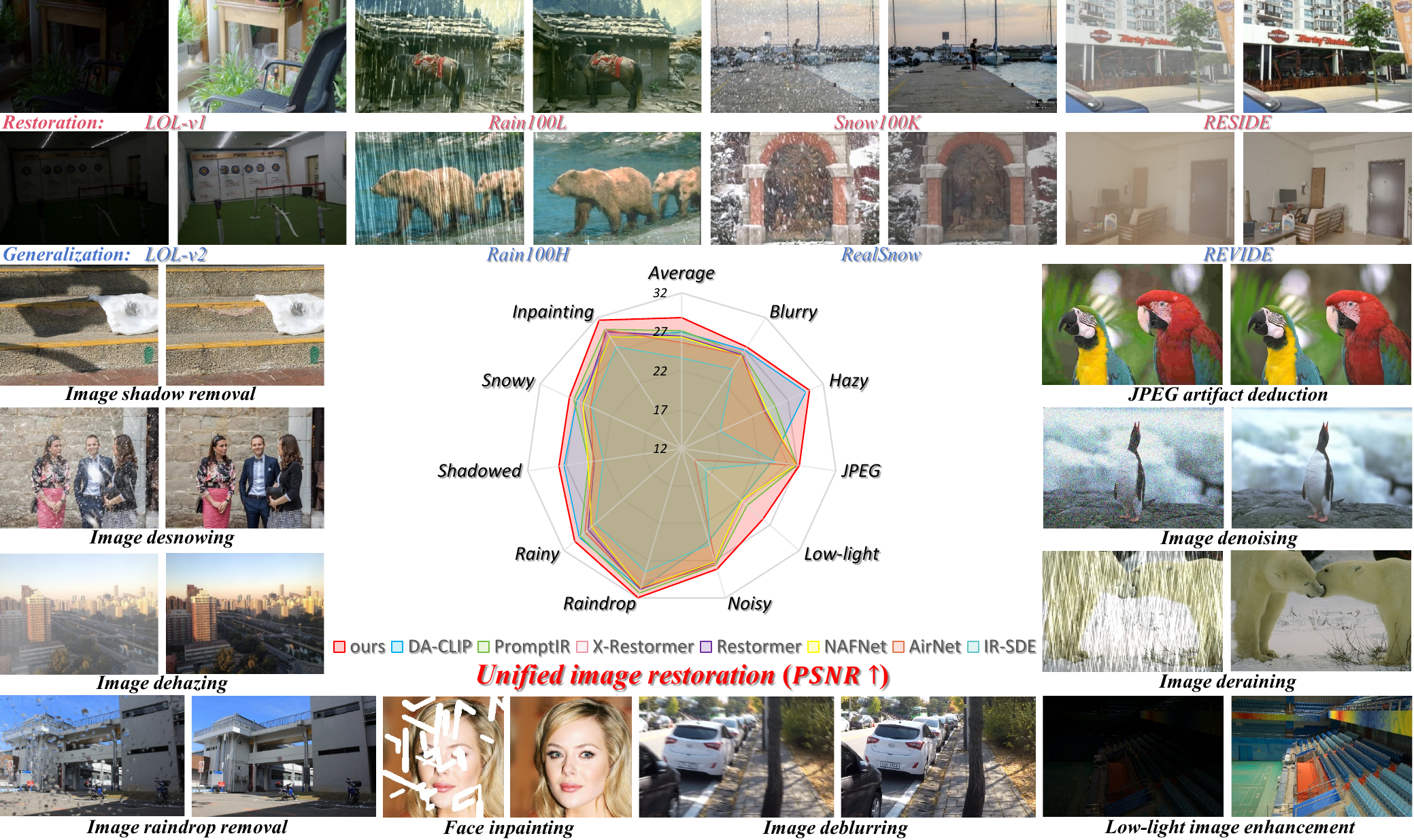}
\vspace{-1.5mm}
   \caption{
Visualizations of our method on image restoration (IR) tasks. When handling out-of-distribution data in single-task IR, our method maintains excellent generalization. For multi-task unified IR, our approach achieves greater performance across 10 degradations.}
   \label{fig:rader1}
\end{figure}
\vspace{-2.9mm}

\begin{abstract}
\vspace{-2.8mm}
While diffusion models demonstrate strong generative capabilities in image restoration (IR) tasks, their complex architectures and iterative processes limit their practical application compared to mainstream reconstruction-based general ordinary IR networks.
Existing approaches primarily focus on optimizing network architecture and diffusion paths but overlook the integration of the diffusion training paradigm within general ordinary IR frameworks.
To address these challenges, this paper elucidates key principles for adapting the diffusion training paradigm to general IR training through systematic analysis of time-step dependencies, network hierarchies, noise-level relationships, and multi-restoration task correlations, proposing a new IR framework supported by diffusion-based training. 
To enable IR networks to simultaneously restore images and model generative representations, we introduce a series of regularization strategies that align diffusion objectives with IR tasks, improving generalization in single-task scenarios.
Furthermore, recognizing that diffusion-based generation exerts varying influences across different IR tasks, we develop an incremental training paradigm and task-specific adaptors, further enhancing performance in multi-task unified IR.
Experiments demonstrate that our method significantly improves the generalization of IR networks in single-task IR and achieves superior performance in multi-task unified IR. Notably, the proposed framework can be seamlessly integrated into existing general IR architectures.
 
\end{abstract}
\vspace{-2.6mm}

\section{Introduction}
\label{sec:intro}

\label{sec:Related Work}
\vspace*{-0.5em}

Image restoration (IR) \cite{Xiao2023RandomST} aims to reconstruct high-quality images from degraded counterparts \cite{Dong2020MultiScaleBD,Nah2021CleanIA,Ren2021AdaptiveCP,Ren2016SingleID,Tsai2021BANetAB,CycleISP,Deblurring9156306,Lu_2024_CVPR,diffu_2025_lu,Lu_2025_CVPR,EvenFormer_2025_lu}. 
Current state-of-the-art approaches primarily employ deep learning and fall into two categories: (1) end-to-end IR networks trained using reconstruction-based objectives \cite{Zhu2023LearningWA,yang2023language,qin2024restoremasksleveragingmask,task9711014}, and (2) image generation models optimized through generation-based objectives \cite{luo2024controlling,luo2023image,AutoDIR,DiffIR}. Comparatively, reconstruction-based methods typically achieve superior fidelity \cite{Zhu2023LearningWA}, whereas generation-based approaches demonstrate enhanced perceptual quality \cite{luo2024controlling}.

Recent generation-based IR methods increasingly utilize diffusion models~\cite{sohl2015deep,ho2020denoising,song2019generative,song2020improved,song2021denoising,song2021maximum,song2021score,rombach2022high,rissanen2022generative,AutoDIR,DiffIR}. Current approaches typically either apply diffusion models directly~\cite{DiffUIR,DreamClean} or fine-tune pre-trained versions for IR tasks~\cite{SUPIR}, though their reliance on large architectures and multi-step sampling limits practical deployment. 
Alternatively, emerging solutions integrate diffusion mechanisms into reconstruction-based training objectives~\cite{tan2024diffloss,liao2024denoising} to enhance generalization. 
Key challenges persist in understanding how diffusion's generative dynamics affect reconstruction-based network training and effectively merging these capabilities for improved general IR performance.

\begin{figure*}[t] 
    \centering
    \includegraphics[width=1\linewidth]{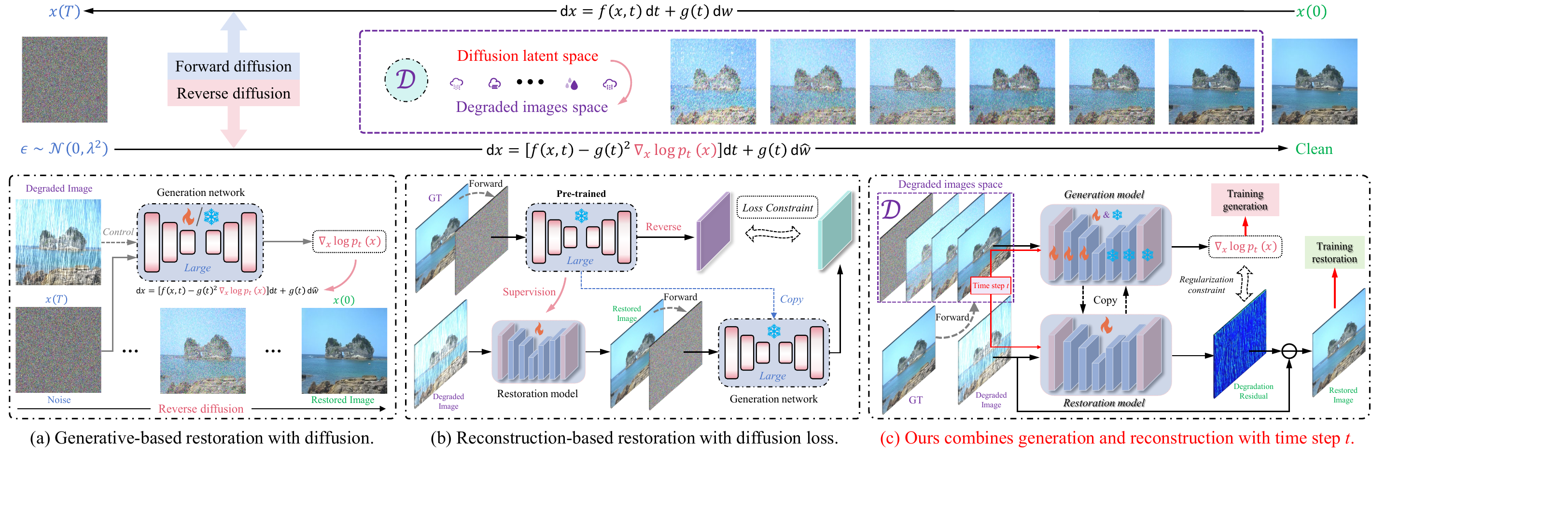}
    \vspace*{-1.5em}
    \caption{
    The top half of the figure illustrates the diffusion generation process, while the bottom half compares our approach with previous methods that employ the diffusion model for image restoration. (a) Generative-based restoration method with diffusion takes the diffusion model as the backbone itself, which produces clean images through reverse diffusion~\cite{DiffIR,DiffUIR}. (b) Reconstruction-based restoration method with diffusion employs typical networks as backbone, taking the latent from a pre-trained diffusion model as an additional loss~\cite{liao2024denoising,tan2024diffloss}. (c) We integrate the diffusion training paradigm into general IR networks, combining generation and reconstruction training.
    }
    \vspace*{-0.8em}
    \label{fig:diffu}
\end{figure*}

In this study, we integrate the diffusion training paradigm into the training of general reconstruction-based IR networks, investigating its impact on performance across various IR tasks. 
Unlike previous work that used diffusion models only as a loss component, we connect the generation training objective of diffusion with the reconstruction training objective of IR networks (see \fref{fig:diffu}). This approach allows us to analyze the influence of the diffusion training mechanism on reconstruction-based IR from multiple perspectives, including time-step dependencies, network hierarchies, noise-level relationships, and multi-restoration task correlations. Detailed exploratory experimental analyses are presented in \sref{Analysis}. 
Our experiments demonstrate that pre-training with diffusion generative objectives enhances the generalization capabilities of IR networks. 
For single-task IR, selecting appropriate time steps, network layers, and the mixing ratio of diffusion-based training can significantly improve model performance. 
For multi-task unified IR, specific time steps help in distinguishing degradation-specific features, and data incremental training guided by these time steps further enhances overall IR efficacy.

We consequently developed distinct methodologies for two application scenarios (see Fig. \ref{fig:pipeline}). For single-task IR, our approach combines diffusion-based pre-training with adaptive fine-tuning guided by time-step-derived prompts. This framework implements specialized regularization strategies that preserve generative capabilities in selected network layers while maintaining reconstruction objectives, ultimately enhancing model generalization. For multi-task unified IR, we build upon generative pre-training and propose an incremental training strategy guided by degradation-specific matching time steps. This approach progressively integrates data from different IR tasks into the training process. Additionally, we introduce MoE adapters for time-step-derived prompts, tailored to specific degradations, to further enhance multi-task IR performance. The visual performance of our method is shown in Fig. \ref{fig:rader1}.
In summary, our contributions are listed as follows:

\setlength{\leftmargini}{2em}
\begin{itemize}

\item
We first comprehensively explore the role of diffusion generation mechanism from a training view in promoting the performance of general ordinary IR networks, analyzing the impact of introducing additional diffusion training paradigm on the main reconstruction-based IR from multiple perspectives, including time-step dependencies, network hierarchies, noise-level relationships, and multi-restoration task correlations.

\item 
We propose a new diffusion training enhanced IR framework. By combining the diffusion generation objective with the main IR network reconstruction objective, this framework implements specialized regularization strategies that preserve generative capabilities in selected network layers while maintaining reconstruction objectives, ultimately enhancing model generalization.

\item 
For single-task IR, we propose utilizing diffusion-based generative pre-training and adaptive fine-tuning based on degradation-specific matching time step $t$ to enhance performance and generalization effects of general reconstruction-based IR networks.

\item 
For multi-task unified IR, we propose an incremental training strategy guided by degradation-specific matching time steps and introduce MoE adapters with time-based prompts to further enhance the multi-task performance of general reconstruction-based IR networks.
\end{itemize}
Extensive experiments demonstrate that our approach effectively enhances the generalization for single-task IR and the stability for multi-task unified IR. Additionally, our method shows good extensibility to existing IR backbones.

\section{Analysis and Methodology Design}
\label{Analysis}
\vspace*{-0.5em}
Recent advances leverage stochastic differential equations (SDEs) based diffusion to progressively map images to Gaussian distributions~\cite{song2021maximum,song2021score,yang2022diffusion}. 
Let \( p_0 \) denote the data's initial distribution, and \( t \in [0, T] \) is a continuous time variable. A diffusion process $\{{x}(t)\}_{t=0}^T$ is defined by following SDE:
\begin{equation}
	\diff {x} = f({x}, t) \diff t + g(t)\diff w, \quad {x}(0) \sim p_0({x}), 
	\label{equ:sde}
\end{equation}
where $f$ and $g$ are the drift and dispersion functions, respectively, $w$ is a standard Wiener process, and ${x}(0) \in \R^{d}$ is an initial condition. We can reverse the process by sampling data from noise through reverse-time simulation of the SDE~\cite{song2021score}, which is defined as follows\cite{anderson1982reverse}:
\begin{equation}
    \begin{split}
        \diff {x} &= \Bigl[ f({x}, t) - g(t)^2\, \nabla_{{x}} \log p_t({x}) \Bigr] \diff t + g(t) \diff \hat{w},
    \end{split}
    \label{equ:reverse-sde}
\end{equation}
where ${x}(T) \sim p_T({x})$, follows a Gaussian distribution, $\hat{w}$ is a reverse-time Wiener process, and $p_t({x})$ is the marginal probability density function of ${x}(t)$ at time $t$. The score function $\nabla_{x} \log p_t({x})$ 
is often addressed by training a time-dependent neural network $s_\theta({x}, t)$ using a score matching objective~\citep{hyvarinen2005estimation,song2021score}.
\fref{fig:diffu} illustrates the forward modeling and reverse solving process of the diffusion model.
Considering that the diffusion latent space inherently contains numerous mappings from degraded to clean images, why not integrate this prior directly with general IR networks?
\fref{fig:diffu}(c) illustrates our diffusion in general IR networks, distinct from the previous methods in \fref{fig:diffu}(a-b).

\subsection{Generative Pre-training and Fine-tuning}
\begin{figure*}[t] 
    \centering
    \includegraphics[width=.98\linewidth]{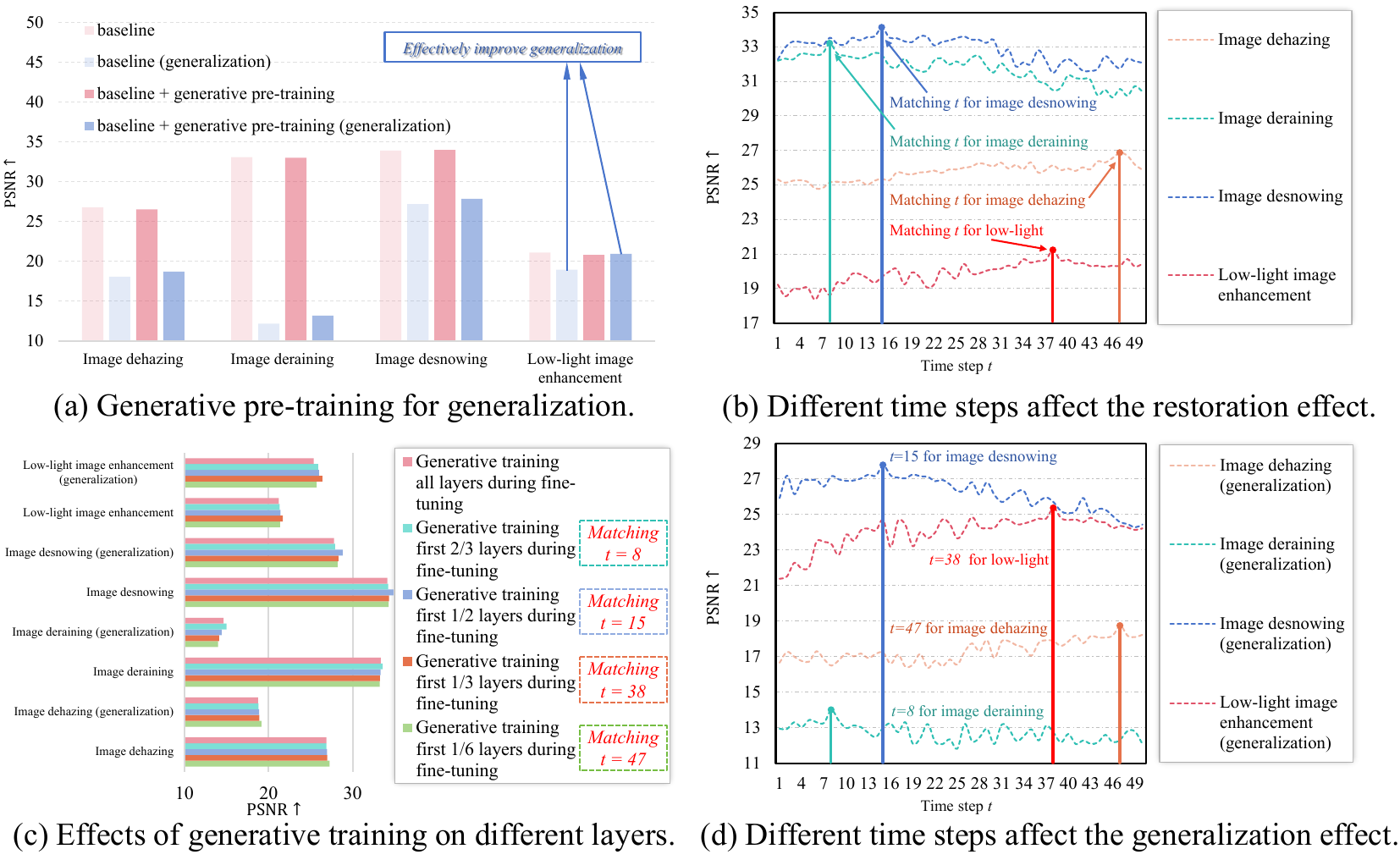}
    \vspace*{-0.5em}
    \caption{
    \textbf{Generative pre-training and fine-tuning} the pre-trained IR network with generative and reconstruction objectives (see Fig.~\ref{fig:diffu} (c)). 
    For generalization, dehazing from RESIDE~\cite{OTS2019} to REVIDE~\cite{9578789}, deraining from Rain100L~\cite{Rain100H_2017_CVPR} to Rain100H~\cite{Rain100H_2017_CVPR}, desnowing from Snow100K~\cite{Snow100K2018} to RealSnow~\cite{Zhu2023LearningWA}, low-light enhancement from LOL-v1~\cite{wei2018deep} to LOL-v2~\cite{Yang2021SparseGR}.
    (a) The generative pre-training empowers the IR model with good generalization.
    (b) Optimal restoration effectiveness is achieved by selecting a time step $t$ matching each degradation type. 
    (c) Generative training that optimizes shallow network layers during fine-tuning enhances performance. The optimal network layers required for achieving the best restoration and generalization effects vary across different degradation types, decreasing as the matching embedded time step $t$ increases.
    (d) Setting different matching $t$ for different IR tasks also ensures optimal generalization.
    }
    \vspace*{-1.8em}
    \label{fig:time}
\end{figure*}

\begin{figure}[tbp]
\begin{minipage}[t]{0.59\columnwidth}
\vspace{0cm}
\setlength{\abovecaptionskip}{0.1cm}
\setlength{\belowcaptionskip}{0.05cm}
\centering
    \includegraphics[width=.99\linewidth]{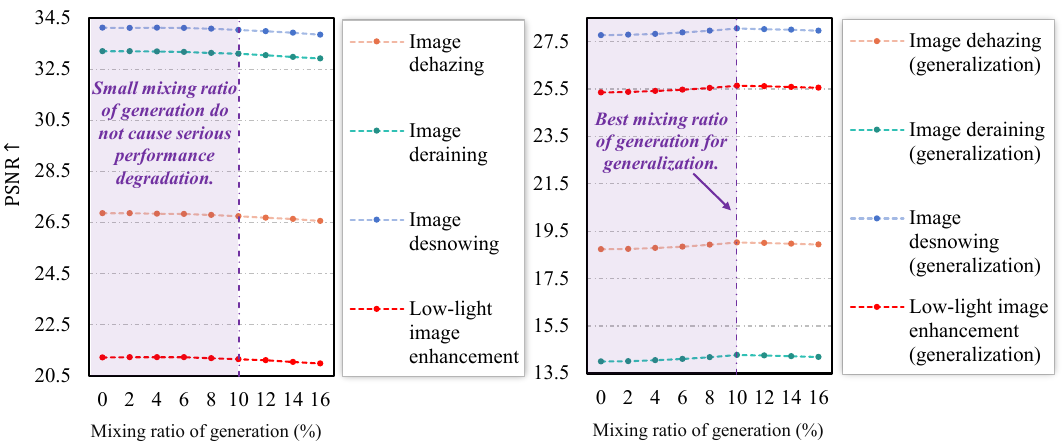}
    \vspace{-1.2mm}
    \caption{
    Mixing ratio of data for generation and reconstruction during fine-tuning
    affect restoration and generalization.
    A suitable mixing ratio does not result in severe performance degradation and further enhances generalization.
    }  
    \label{mix}
\end{minipage}
\hfill
\begin{minipage}[t]{0.39\columnwidth}
\vspace{0mm}
\setlength{\abovecaptionskip}{0.01cm}
\setlength{\belowcaptionskip}{0.05cm}
\centering
    \includegraphics[width=.99\linewidth]{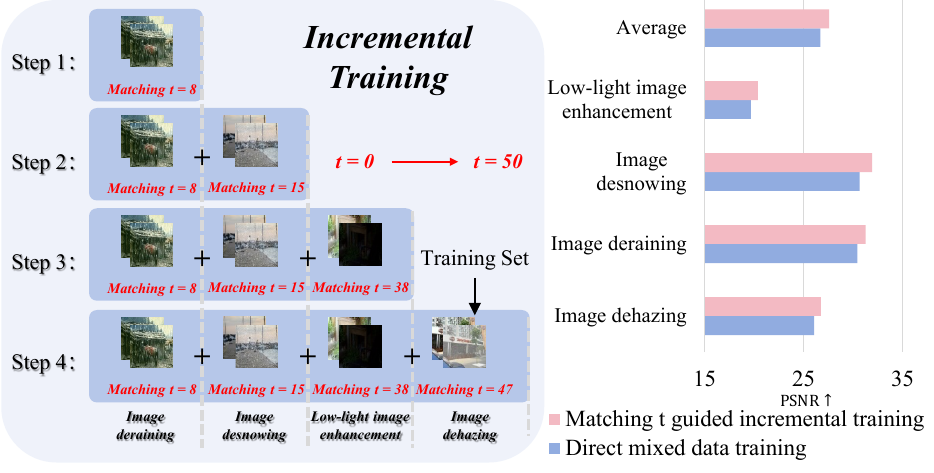}
    \vspace{-1.2mm}
    \caption{Injecting the training data incrementally in the order of diffusion time steps, which is derived from the different task data's matching $t$, leading to a better unified IR network.}  
    \label{fig:incremental}
\end{minipage}
\vspace{-1.8em}
\end{figure}

We introduce the commonly used generative pre-training~\cite{2018Improving} and task-specific fine-tuning techniques from the image generation field into image restoration, as implemented in a U-Net~\cite{luo2023image}. 
By incorporating generative pre-training into 4 IR tasks, we fine-tune the obtained networks for image restoration. \fref{fig:time} and \fref{mix} analyze the experimental results in detail.
\fref{fig:time}(a) first shows that \textbf{generative pre-training can significantly improve the generalization effect on single-task IR.}

\noindent
\textbf{Degradation-specific matching time step enhances fine-tuning.} 
Due to integrating diffusion's generative training objective, the IR network incorporates a time step parameter, \textit{i.e.}, $r_\theta({x}, t)$, where \( t \in [0, T] \), and $\theta$ is the parameters of network $r$.
Unlike setting $T$ to 1000 in image generation, we set $T$ to 50 like DDIM~\cite{DDIM}. This choice increases the noise intensity in predicting each time step during generative training, better coupling with the predicted degradation residue in IR tasks, and reducing training complexity.
We further analyze the impact of $t$ on IR network during fine-tuning, shown in \fref{fig:time}(b) and \fref{fig:time}(d). It is observed that for various IR tasks, the optimal select of $t$ is different, which enhances both source domain IR performance and improves generalization effects.
By performing the diffusion and reverse diffusion process for degraded image on the pre-trained genenative model ${{r_\theta}_{pre}}$ with different time $t$, we can easily obtain the matching $t^{mat}$, which effectively aligns the reversed image with the clean image.
Experimental results show that the size of $t^{mat}$ to some extent characterizes the complexity of the degradation type. The increase of $t^{mat}$ corresponds to more global and complex degradation situations, such as low light and haze, which is similar to the progressive distribution of noise in the diffusion process.
Drawing inspiration from the forward process of diffusion models, we incrementally inject different degradation data into the model in order of their matching time step $t^{mat}$, resulting in a unified IR network. 
As shown in \fref{fig:incremental}, our incremental training based on $t^{mat}$ enhances the network's unified IR performance effectively.

\noindent
\textbf{Mix additional data for generation training objectives during the main reconstructive IR fine-tuning.}
Training with mixed data is commonly used in image generation~\cite{song2019generative}. We attempt to incorporate additional generative training into the specific degradation IR fine-tuning. 
\fref{mix} illustrates that mix generative training during IR task fine-tuning can further stabilize the generalization effects. 
Increasing the mix ratio of data for the diffusion's generative training objective to around 10\% reaches a bottleneck in improvement, further increasing the generative ratio impairs the IR effectiveness.

\subsection{Generation and Reconstruction Gaps}
The analysis above can provide insights into three feasible applications of diffusion in general IR networks: \textbf{(1)} Generative pre-training can enhance the generalization performance of IR networks. \textbf{(2)} The degradation-specific matching time step $t$ obtained from generative pre-training can align with different degradation types, improving the IR effectiveness after fine-tuning. \textbf{(3)} Mixing an appropriate ratio of generative training in fine-tuning can further enhance the IR outcomes.
While diffusion can effectively enhance IR, the gaps between generative and reconstructive training remains unresolved. Below, we explain the reasons for these gaps and provide corresponding solutions.

\noindent
\textbf{Point 1: Gap of generation and reconstruction objectives affects IR performance.}
Generative pre-training benefits the generalization of IR models, but as task-specific fine-tuning progresses, catastrophic forgetting \cite{McCloskey1989CatastrophicII} cause this generalization ability to diminish. Additionally, the mix of generative and reconstructive training in fine-tuning may face challenges in achieving optimal results due to differences in gradient optimization directions. Therefore, constraining the gaps of generation and reconstruction objectives is necessary to ensure generalization and restoration performance.

\noindent
\textbf{Solution 1: Parameter importance regularization and gradient orthogonality techniques constrain the gap.}
Inspired by continual learning \cite{EWC2017}, we utilize a parameter regularization strategy with second-order parameter importance modification schemes.
We first calculate the parameter importance of the IR network obtained by generative pre-training, then update the network parameters according to the importance weights in fine-tuning. 
Specifically, denoting pre-training as task 0, fine-tuning as task 1, parameter importance weights $\Omega_{\theta_{k}}$  are calculated by accumulating gradients:
\begin{equation}
    \begin{split}        \Omega_{\theta_{k}}=f(x;\theta_{r}^{1})-f(x;\theta_{k}^{0}),
    \end{split}
    \label{importance1}
\end{equation}
where $f(\cdot)$ represents the mapping function of IR network, $\theta_{k}$ denotes $k$th parameter of the IR network, and $\delta\theta_{k}^{1}=\delta\theta_{k}^{0}+\delta\theta_{k}$. $\delta$ denotes the parameter change magnitude, and indicates the input various degradation data. In particular, the above equation can be written as:
\vspace*{-0.5em}
\begin{equation}
    \begin{split}        \Omega_{\theta_{k}}=\nabla_{\theta_{k}}\mathcal{L}\,|\delta\theta_{k}|+\frac{1}{2}\cdot\nabla_{\theta_{k}}^{2}\mathcal{L}\,|\delta\theta_{k}|^{2}+O(|\delta\theta_{k}|^{3}),
    \end{split}
    \label{importance2}
\end{equation}
where $\mathcal{L}$ is the conventional loss of baseline method. 
We retain the first two terms as second-order importance and obtain the parameter regularization loss that maintains the generalization ability of generative pre-training in restoration fine-tuning as follows:
\begin{equation}
    \begin{split}        
    {\mathcal{L}_{reg}}=\lambda\sum_{k=1}^{m}\left[\nabla_{\theta_{k}}{\mathcal{L}}\,|\delta\theta_{k}|+{\frac{1}{2}}\cdot\nabla_{\theta_{k}}^{2}\,{\mathcal{L}}\,|\delta\theta_{k}|^{2}\right],
    \end{split}
    \label{importance}
\end{equation}
where $\lambda$ is balance coefficient, $m$ is number of parameters.

During mixed generative task's restoration fine-tuning process, we incorporate gradient orthogonality to further ensure training efficiency.
Given input-gt pairs $\{{x}_i,y_i\}_{i=0}^B$, generative training with $i \in {B_g}$ and reconstruction training with $i \in {B_r}$. Let $\mathbf{g}_i=\nabla_\theta\mathcal{L}=\partial \mathcal{L}/\partial \theta$ be the gradient vector during fine-tuning. 
We design a unified gradient orthogonal loss $\mathcal{L}_{orthog}$ to constrain the gradient direction of generation and reconstruction during fine-tuning as follows:
\begin{equation}
    \begin{split}        
s = \sum\limits_{i \in {B_g},j \in {B_r}} {\left\langle {{\mathbf{g}_i},{\mathbf{g}_j}} \right\rangle /\sum\limits_{i \in {B_g},j \in {B_r}} 1 },
    \end{split}
    \label{s}
\end{equation}
\begin{equation}
    \begin{split}        
d = \left( {\sum\limits_{i,k \in {B_g}} {\left\langle {{\mathbf{g}_i},{\mathbf{g}_k}} \right\rangle }  + \sum\limits_{l,j \in {B_r}} {\left\langle {{\mathbf{g}_l},{\mathbf{g}_j}} \right\rangle } } \right)/\sum\limits_{i,k,l,j \in {B}} 1 ,
    \end{split}
    \label{d}
\end{equation}
\begin{equation}
    \begin{split}        
{\mathcal{L}_{orthog}} = \left( {1 - s} \right) + \left| d \right|,
    \end{split}
    \label{zjloss}
\end{equation}
 where $\left\langle { \cdot , \cdot } \right\rangle$ is the cosine similarity operator applied on two
 vectors, $\left|  \cdot  \right|$ is the absolute value operator, and $B$ is
 mini-batch size. Note that the cosine similarity operator 
 used in Eq. \ref{s} and Eq. \ref{d} involves normalization of features (projection to a unit hyper-sphere) as follows:
\begin{equation}
    \begin{split}        
\left\langle {{\mathbf{g}_i},{\mathbf{g}_j}} \right\rangle  = \frac{{{\mathbf{g}_i} \cdot {\mathbf{g}_j}}}{{{{\left\| {{\mathbf{g}_i}} \right\|}_2} \cdot {{\left\| {{\mathbf{g}_j}} \right\|}_2}}},
    \end{split}
    \label{cos}
\end{equation}
where $\left\|  \cdot  \right\|_2$ refers to the $\ell_2$ norm operator.

\noindent
\textbf{Point 2: Gap of generation needs across different layers varies in IR network.}
In the inference process of a well-trained IR network, as illustrated in \fref{fig:layer_feature_real}, a comparison between shallow and deep feature maps reveals that shallow network layers retain more background information (\textit{\textbf{domain-invariant}}), which is crucial for generalization. In contrast, deep layers focus on the residuals of specific degradations (\textit{\textbf{domain-specific}}), which impacts the performance ceiling of the IR network for specific tasks. 
By ensuring that gradient updates from mixed generative training during fine-tuning only affect shallow network layers, as depicted in \fref{fig:time}(c), we can enhance source domain IR performance while maintaining generalization, validating our viewpoint. Therefore, different network layers have varying needs for diffusion generative training, and it is essential to allocate the layers for generative training affects the fine-tuning process judiciously.

\noindent
\textbf{Solution 2: Constrain gaps by generation-sensitive layers.}
Based on the results of \fref{fig:time}(c), for mixed generative and reconstructive training during fine-tuning, we perform gradient updates only in the shallow layers which sensitive to generative objectives but less affect reconstructive objectives and thus constrain the gap, applying weight decay to the parameters with matching time step $t^{mat}$: 
\begin{equation}
    \begin{split}        
w_{decay}(t) = e^{-at} ,
    \end{split}
    \label{decay}
\end{equation}
where a is the decay rate, which is set to 0.05.
During the reconstructive training in the fine-tuning process, we also adjust the balancing coefficient $\lambda$  for the regularization loss $\mathcal{L}_{reg}$ in Eq. \ref{importance} based on the layer-wise decay of the network when applying parameter importance regularization.

\begin{figure*}[t] 
    \centering
    \includegraphics[width=1\linewidth]{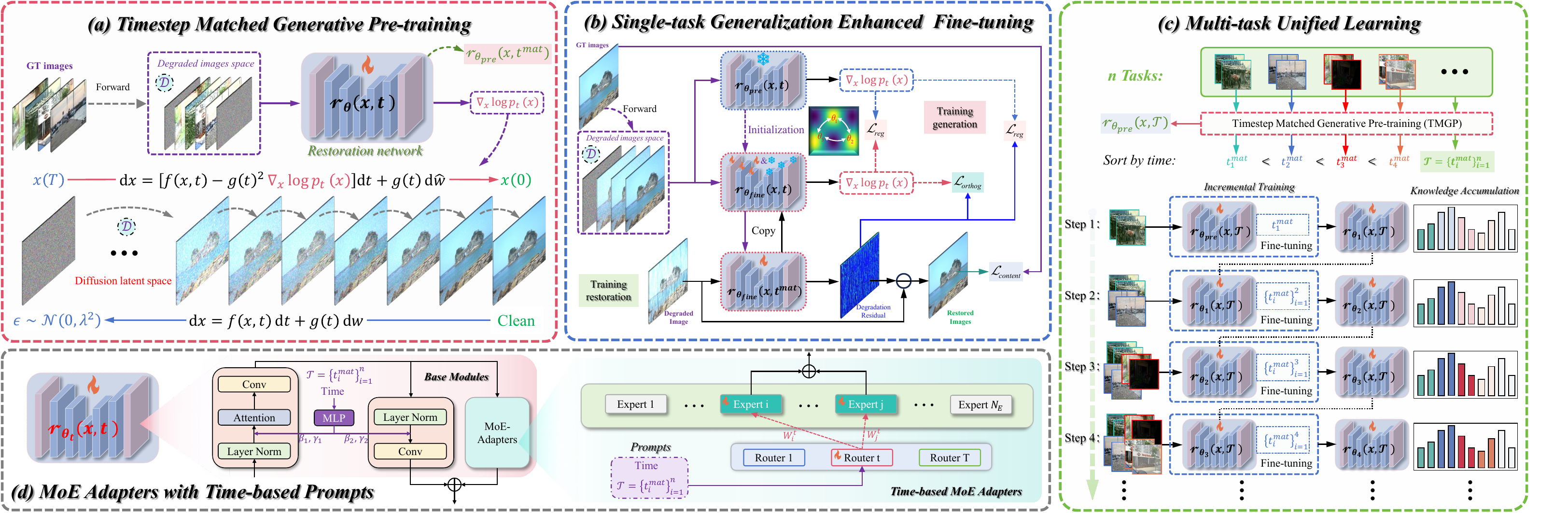}
    \vspace*{-1.5em}
    \caption{
    Overall schema of our proposed method with diffusion's generative training mechanism.    
    (a) Timestep Matched Generative Pre-training (TMGP). 
    We use GT images to pre-train the IR network with diffusion's training objective to obtain a generative model $r_{\theta_{pre}}$, and use reverse diffusion to obtain the matching time step $\mathcal{T}={\{t_i^{mat}\}}_{i=1}^n$ that best matches specific degradation types.
    (b) Generalization Enhanced Fine-tuning (GEF).
    We perform image restoration training on model $r_{\theta_{pre}}$ with generative and reconstructive training objectives, taking regularization loss $\mathcal{L}_{reg}$ that uses parameter importance to guide updates to avoid the model forgetting previously generative knowledge, and use gradient orthogonality loss $\mathcal{L}_{orthog}$ to enhance generalization capabilities by reducing the conflicts between generative and reconstructive optimization.
    (c) Multi-task Unified Learning (MTUL).
    After generative pre-training, the network gradually learns and accumulates knowledge through Time-sequential Incremental Training (TIT) to develop a unified IR model.
    }
    \vspace*{-1.5em}
    \label{fig:pipeline}
\end{figure*}

\section{Diffusion Training Enhanced IR Framework}
\label{method}
\vspace*{-0.5em}
Based on the analysis and method mechanism in \sref{Analysis}, we proposed a universal IR framework based on diffusion's generative training in \fref{fig:pipeline}, which can further improve the generalization in a single task and the unified learning ability of multiple IR tasks for general IR networks.
\subsection{Timestep Matched Generative Pre-training}
As shown in \fref{fig:pipeline}(a), model $r_{\theta_{pre}}$ learns the prior knowledge in diffusion latent space through generative pre-training on GT images.
Similar to DDPM~\cite{ho2020denoising}, our optimization objective is:
\begin{equation}
    \mathcal{L}_{\gamma}(\theta) \coloneqq \sum_{t=1}^T \gamma_t \, \mathbb{E} \Big[ \big\lVert \tilde{r}_\theta({x}_{t}, x_0, t) - \epsilon_{t} \bigr\rVert\Big],
    \label{eq:noise_objective}
\end{equation}
where $\gamma_1, \ldots, \gamma_T$ are loss weights.
By applying diffusion and reverse diffusion process for degradation image with different time $t$, we can find $t_i^{mat}$ that best matches reversed image with clean image, which helps us distinguish different degradation types and match them to the latent space of diffusion.

\subsection{Generalization Enhanced Fine-tuning}
As shown in \fref{fig:pipeline}(b), model $r_{\theta_{fine}}$ learns the degradation knowledge during Generalization Enhanced Fine-tuning (GEF).  
Parameter importance guided updates in \eref{importance2} avoid the model forgetting previously generative knowledge, gradient orthogonality in \eref{zjloss} enhances generalization capabilities through mixed generation training. 
Total optimization objective for GEF is defined as follows:
\begin{equation}    \mathcal{L}_{fine}=\mathcal{L}_{content}+\mathcal{L}_{reg}+\mathcal{L}_{orthog},
    \label{eq:fine_objective}
\end{equation}
where $\mathcal{L}_{content}$ is $\ell_1$ loss used in image restoration.

\subsection{Effective Multi-task Unified Learning}
As shown in \fref{fig:pipeline}(c), model $r_{\theta_{T}}(x,\mathcal{T})$ accumulates multi-task knowledge through Time-sequential Incremental Training (TIT). 
\fref{fig:pipeline}(d) indicates Our MoE Adapters with Time-based Prompts.

\begin{table*}[!t]
\begin{center}
\caption{\small \textbf{{{Single-task IR and generalization}}} (\textcolor{blue! 100}{Setting A}) comparison. Our generalization is optimal.
} 
\label{table:genera_4}
\vspace*{-2mm}
\setlength{\tabcolsep}{2.7pt}
\scalebox{0.7}{
    \begin{tabular}{c|c|c|c|c|c||c|c|c|c|c|c}
    \toprule [0.15em]
    \textcolor{blue! 100}{Image dehazing}  & \multirow{2}[4]{*}{Venue} & \multicolumn{2}{c|}{Restoration} & \multicolumn{2}{c||}{Generalization} & \textcolor{blue! 100}{Image Deraining}  & \multirow{2}[4]{*}{Venue} & \multicolumn{2}{c|}{Restoration} & \multicolumn{2}{c}{Generalization} \\
\cmidrule{1-1}\cmidrule{3-7}\cmidrule{9-12}    Method &       & PSNR↑ & SSIM↑ & PSNR↑ & SSIM↑ & Method &       & PSNR↑ & SSIM↑ & PSNR↑ & SSIM↑ \\
    \midrule
    GridDehazeNet~\citep{liu2019griddehazenet} & ICCV 19 & 25.86 & 0.944 & 16.53 & 0.619 & PReNet~\citep{ren2019progressive} & CVPR 19 & 37.48  & 0.979 & 13.04 & 0.508 \\
    MAXIM~\citep{tu2022maxim} & CVPR 22 & 29.12 & 0.932 & 17.01 & 0.653 & MPRNet~\cite{zamir2021multi} & CVPR 21 & 36.40  & 0.965 & 13.25 & 0.514 \\
    DehazeFormer~\citep{song2023vision}  & TIP 23 & \textbf{30.29} & \textbf{0.964} & 15.82 & 0.537 & MAXIM~\citep{tu2022maxim} & CVPR 22 & 38.06  & 0.977 & 13.67 & 0.566 \\
    IR-SDE~\cite{luo2023image} & ICML 23 & 25.25 & 0.906 & 16.63 & 0.624 & IR-SDE~\cite{luo2023image} & ICML 23 & \textbf{38.30 } & \textbf{0.981} & 13.23 & 0.554 \\
    DA-CLIP~\cite{DA-CLIP} & ICLR 24 & 30.16 & 0.936 & 16.38 & 0.636 & RAM~\cite{qin2024restoremasksleveragingmask}   & ECCV 24 & 37.21  & 0.934 & 14.12 & 0.572 \\
    X-Restormer~\cite{X-Restormer}  & ECCV 24 & 29.34 & 0.921 & 17.26 & 0.688 & X-Restormer~\cite{X-Restormer}  & ECCV 24 & 37.69  & 0.968 & 13.45 & 0.529 \\
    Ours  & -     & 29.08 & 0.933 & \textbf{20.74} & \textbf{0.782} & Ours  & -     & 37.47 & 0.971 & \textbf{16.87} & \textbf{0.695} \\
    \midrule
    \midrule
    \textcolor{blue! 100}{Image desnowing} & \multirow{2}[4]{*}{Venue} & \multicolumn{2}{c|}{Restoration} & \multicolumn{2}{c||}{Generalization} & \textcolor{blue! 100}{Low-light enhancement} & \multirow{2}[4]{*}{Venue} & \multicolumn{2}{c|}{Restoration} & \multicolumn{2}{c}{Generalization} \\
\cmidrule{1-1}\cmidrule{3-7}\cmidrule{9-12}    Method &       & PSNR↑ & SSIM↑ & PSNR↑ & SSIM↑ & Method &       & PSNR↑ & SSIM↑ & PSNR↑ & SSIM↑ \\
    \midrule
    NAFNet~\cite{NAF} & ECCV 22 & 34.82 & 0.936 & 26.15 & 0.784  & MIRNet~\citep{zamir2020learning} & ECCV 20 & \textbf{24.14} & 0.830  & 23.97 & 0.815 \\
    MAXIM~\citep{tu2022maxim} & CVPR 22 & 34.93 & 0.944 & 25.85 & 0.762  & EnlightenGAN~\citep{jiang2021enlightengan} & TIP 21 & 17.61 & 0.653  & 19.74 & 0.665 \\
    Restormer~\cite{zamir2022restormer}  & CVPR 22 & \textbf{35.21} & \textbf{0.957} & 25.63 & 0.693  & URetinex-Net~\citep{wu2022uretinex} & CVPR 22 & 19.84 & 0.824  & 20.24 & 0.755 \\
    IR-SDE~\cite{luo2023image} & ICML 23 & 33.84 & 0.955 & 25.95 & 0.730  & IR-SDE~\cite{luo2023image} & ICML 23 & 20.45 & 0.787  & 22.73 & 0.794 \\
    DA-CLIP~\cite{DA-CLIP} & ICLR 24 & 35.16 & 0.931 & 26.03 & 0.747  & DA-CLIP~\cite{DA-CLIP} & ICLR 24 & 23.77 & 0.830  & 23.09 & 0.814  \\
    X-Restormer~\cite{X-Restormer} & ECCV 24 & 35.22 & 0.954 & 25.77 & 0.722  & X-Restormer~\cite{X-Restormer}  & ECCV 24 & 19.25 & 0.811  & 21.25 & 0.796 \\
    Ours  & -     & 34.77 & 0.943 & \textbf{27.84} & \textbf{0.829 } & Ours  & -     & 23.59 & \textbf{0.856} & \textbf{25.67} & \textbf{0.883} \\

    \bottomrule[0.15em]
    \end{tabular}}
\end{center}\vspace{-2em}
\end{table*}

\begin{table*}[!t]
\begin{center}
\caption{\small {\textbf{Multi-task unified IR}} (\textcolor{blue! 100}{Setting B}) comparison. \textbf{Best} and \underline{second best} performances are highlighted.
Compared to state-of-the-art methods, our method has significant improvements in both fidelity and perception indicators. At the same time, extending our solution to the existing advanced restoration networks NAFNet~\cite{NAF} and X-Restormer~\cite{X-Restormer} can further improve the model performance.
} 
\label{table:deg10}
\vspace*{-2mm}
\setlength{\tabcolsep}{3.2pt}
\scalebox{0.7}{
    \begin{tabular}{c|c|c|c|c|c|c|c|c|c|c|c|c}
    \toprule[0.15em]
    \rowcolor{blue! 6} \textbf{\textcolor{blue! 100}{PSNR↑}} & Venue & Blurry & Hazy  & JPEG  & Low-light & Noisy & Raindrop & Rainy & Shadowed & Snowy & Inpainting & Average \\

    \midrule
    AirNet~\cite{AirNet} & CVPR 22 & 26.25  & 23.56  & 26.98  & 14.24  & 27.51  & 30.68  & 28.45  & 23.48  & 24.87  & 30.15  & 25.62  \\
    Restormer~\cite{zamir2022restormer} & CVPR 22 & 26.34  & 23.75  & 26.90  & 22.17  & 27.25  & 30.85  & 27.91  & 23.33  & 25.98  & 29.88  & 26.44  \\
    NAFNet~\cite{NAF} & ECCV 22 & 26.12  & 24.05  & 26.81  & 22.16  & 27.16  & 30.67  & 27.32  & 24.16  & 25.94  & 29.03  & 26.34  \\
    PromptIR~\cite{PromptIR} & NeurIPS 23 & 26.50  & 25.19  & 26.95  & 23.14  & 27.56  & 31.35  & 29.24  & 24.06  & 27.23  & 30.22  & 27.14  \\
    IR-SDE~\cite{luo2023image} & ICML23 & 24.13  & 17.44  & 24.21  & 16.07  & 24.82  & 28.49  & 26.64  & 22.18  & 24.70  & 27.56  & 23.62  \\
    DA-CLIP~\cite{DA-CLIP} & ICLR 24 & 27.03  & 29.53  & 23.70  & 22.09  & 24.36  & 30.81  & 29.41  & 27.27  & 26.83  & 28.94  & 27.00  \\
    ResShift~\cite{ResShift} & TPAMI 24 & 27.12  & 25.83  & 24.24  & 21.54  & 26.82  & 29.37  & 28.95  & 25.76  & 26.69  & 29.51  & 26.58  \\
    X-Restormer~\cite{X-Restormer} & ECCV 24 & 26.77  & 23.65  & 25.73  & 22.56  & 27.25  & 29.88  & 28.45  & 23.97  & 26.31  & 29.68  & 26.43  \\    RAM~\cite{qin2024restoremasksleveragingmask}   & ECCV 24 & \textbf{27.55 } & 25.56  & \textbf{27.42 } & 23.04  & 27.54  & 29.97  & 29.27  & 26.69  & \underline{27.65}  & 30.38  & 27.51  \\
    \midrule
    Ours  & -     & 27.10  & 29.84  & 27.01  & \underline{23.82}  & 27.53  & 30.93  & 29.88  & \textbf{27.43 } &  27.35  & 30.36  & 28.13  \\
    Ours + NAFNet & -     & 27.46  & \underline{30.06}  & \underline{27.27}  & \textbf{24.72 } & \underline{28.17}  & \underline{31.36}  & \textbf{30.12 } & \underline{27.35}  & 27.57  & \textbf{30.60 } & \textbf{28.47 } \\
    Ours + X-Restormer & -     & \underline{27.50}  & \textbf{30.23 } & 27.11  & 22.98  & \textbf{28.24 } & \textbf{31.59 } & \underline{30.05}  & 27.22  & \textbf{27.84 } & \underline{30.42}  & \underline{28.32}  \\
    \midrule
    \midrule
    \rowcolor{yellow! 8}\textbf{\textcolor{blue! 100}{SSIM↑}} & Venue & Blurry & Hazy  & JPEG  & Low-light & Noisy & Raindrop & Rainy & Shadowed & Snowy & Inpainting & Average \\

    \midrule
    AirNet~\cite{AirNet} & CVPR 22 & 0.805  & 0.916  & 0.783  & 0.781  & 0.769  & 0.926  & 0.867  & 0.832  & 0.846  & 0.911  & 0.844  \\
    Restormer~\cite{zamir2022restormer} & CVPR 22 & 0.811  & 0.915  & 0.781  & 0.815  & 0.762  & 0.928  & 0.862  & 0.836  & 0.877  & 0.912  & 0.850  \\
    NAFNet~\cite{NAF} & ECCV 22 & 0.804  & 0.926  & 0.780  & 0.809  & 0.768  & 0.924  & 0.848  & 0.839  & 0.869  & 0.901  & 0.847  \\
    PromptIR~\cite{PromptIR} & NeurIPS 23 & 0.815  & 0.933  & 0.784  & \textbf{0.829}  & 0.774  & 0.931  & 0.876  & 0.842  & 0.887  & 0.918  & 0.859  \\
    IR-SDE~\cite{luo2023image} & ICML23 & 0.730  & 0.832  & 0.615  & 0.719  & 0.640  & 0.822  & 0.808  & 0.667  & 0.828  & 0.876  & 0.754  \\
    DA-CLIP~\cite{DA-CLIP} & ICLR 24 & 0.810  & 0.931  & 0.532  & 0.796  & 0.579  & 0.882  & 0.854  & 0.811  & 0.854  & 0.894  & 0.794  \\
    ResShift~\cite{ResShift} & TPAMI 24 & 0.808  & 0.926  & 0.579  & 0.723  & 0.624  & 0.871  & 0.839  & 0.829  & 0.833  & 0.884  & 0.792  \\
    X-Restormer~\cite{X-Restormer} & ECCV 24 & 0.805  & 0.911  & 0.774  & 0.806  & \underline{0.789}  & 0.917  & 0.855  & 0.846  & 0.898  & 0.917  & 0.852  \\
    RAM~\cite{qin2024restoremasksleveragingmask}  & ECCV 24 & 0.817  & \textbf{0.894 } & 0.765  & 0.710  & 0.771  & 0.925  & 0.860  & 0.854  & 0.884  & 0.920  & 0.840  \\
    \midrule
    Ours  & -     & 0.814  & 0.883  & 0.796  & \underline{0.818 } & 0.787  & \underline{0.947}  & 0.877  & 0.859  & \textbf{0.904 } & 0.935  & \underline{0.862}  \\
    Ours + NAFNet & -     & \underline{0.821}  & \underline{0.892}  & \textbf{0.804 } & 0.812  & 0.774  & \textbf{0.954 } & \textbf{0.895 } & \underline{0.862}  & \underline{0.897}  & \textbf{0.942 } & \textbf{0.865 } \\
    Ours + X-Restormer & -     & \textbf{0.825 } & {0.887}  & \underline{0.802}  & 0.759  & \textbf{0.798 } & 0.944  & \underline{0.887}  & \textbf{0.865 } & 0.889  & \underline{0.938}  & 0.859  \\
    \midrule
    \midrule
    \rowcolor{red! 8}\textbf{\textcolor{blue! 100}{LPIPS↓}} & Venue & Blurry & Hazy  & JPEG  & Low-light & Noisy & Raindrop & Rainy & Shadowed & Snowy & Inpainting & Average \\

    \midrule
    AirNet~\cite{AirNet} & CVPR 22 & 0.279  & 0.063  & 0.302  & 0.321  & 0.264  & 0.095  & 0.163  & 0.145  & 0.112  & 0.071  & 0.182  \\
    Restormer~\cite{zamir2022restormer} & CVPR 22 & 0.282  & 0.054  & 0.300  & 0.156  & \textbf{0.215}  & 0.083  & 0.170  & 0.145  & 0.095  & 0.072  & 0.157  \\
    NAFNet~\cite{NAF} & ECCV 22 & 0.284  & \underline{0.043}  & 0.303  & 0.158  & \underline{0.216 } & 0.082  & 0.180  & 0.138  & 0.096  & 0.085  & 0.159  \\
    PromptIR~\cite{PromptIR} & NeurIPS 23 & 0.267  & 0.051  & 0.269  & 0.140  & 0.230  & 0.078  & 0.147  & 0.143  & 0.082  & 0.068  & 0.148  \\
    IR-SDE~\cite{luo2023image} & ICML23 & 0.198  & 0.168  & \textbf{0.246 } & 0.185  & 0.232  & 0.113  & 0.142  & 0.223  & 0.107  & 0.065  & 0.168  \\
    DA-CLIP~\cite{DA-CLIP} & ICLR 24 & \underline{0.140}  & \textbf{0.037 } & 0.317  & 0.114  & 0.272  & 0.068  & \textbf{0.085 } & 0.118  & \textbf{0.072 } & \underline{0.047}  & \underline{0.127}  \\
    ResShift~\cite{ResShift} & TPAMI 24 & 0.164  & 0.064  & 0.281  & 0.155  & 0.294  & 0.102  & 0.097  & 0.124  & 0.088  & 0.048  & 0.142  \\
    X-Restormer~\cite{X-Restormer} & ECCV 24 & 0.297  & 0.074  & 0.305  & 0.168  & 0.254  & 0.080  & 0.181  & 0.151  & 0.122  & 0.074  & 0.171  \\
    RAM~\cite{qin2024restoremasksleveragingmask}   & ECCV 24 & 0.284  & 0.079  & 0.298  & 0.148  & 0.269  & 0.085  & 0.169  & 0.174  & 0.110  & 0.102  & 0.172  \\
    \midrule
    Ours  & -     & 0.155  & 0.052  & 0.262  & \underline{0.112}  & 0.252  & 0.066  & 0.097  & \textbf{0.105 } & 0.080  & \textbf{0.046 } & \textbf{0.123 } \\
    Ours + NAFNet & -     & \textbf{0.137 } & 0.066  & 0.292  & 0.117  & 0.259  & \underline{0.065}  & \underline{0.091}  & \underline{0.109}  & 0.082  & 0.051  & \underline{0.127}  \\
    Ours + X-Restormer & -     & 0.161  & 0.054  & \underline{0.258}  & \textbf{0.110 } & 0.277  & \textbf{0.061 } & 0.115  & 0.125  & \underline{0.075}  & 0.062  & 0.130  \\
   
    \bottomrule[0.15em]
    \end{tabular}
    }
\end{center}\vspace{-2.2em}
\end{table*}

\section{Experiments}
\vspace*{-0.5em}
{\textbf{Single-task IR and generalization}} (\textcolor{blue! 100}{Setting A})
: dehazing from RESIDE~\cite{OTS2019} to REVIDE~\cite{9578789}, deraining from Rain100L~\cite{Rain100H_2017_CVPR} to Rain100H~\cite{Rain100H_2017_CVPR}, desnowing from Snow100K~\cite{Snow100K2018} to RealSnow~\cite{Zhu2023LearningWA}, low-light enhancement from LOL-v1~\cite{wei2018deep} to LOL-v2~\cite{Yang2021SparseGR}.

{\textbf{Multi-task unified IR}} (\textcolor{blue! 100}{Setting B}): 10 restoration tasks as shown in \tref{table:dataset}. We use peak signal-to-noise ratio (PSNR) and structural similarity index (SSIM)~\citep{wang2004image} for fidelity evaluation, and the learned perceptual image patch similarity (LPIPS)~\citep{zhang2018unreasonable} for perceptual evaluation. 

\noindent
\textbf{Implementation details.}
We implemented our method on the PyTorch platform using 8 NVIDIA GTX 1080Ti GPUs. The optimizer used is Adam with an exponential decay rate of 0.9. The initial learning rate is set to 5e-5, a cosine annealing strategy is employed for adjustment, and other training settings follow~\cite{DA-CLIP}. The number of MoE experts $N_E=10$, $a=0.05$ and $\lambda=0.2$.

\noindent
\textbf{Comparison in} \textcolor{blue! 100}{Setting A}.
Our method has significantly improved the generalization of the four degradation tasks, and the restoration effect on the training dataset is also competitive. The specific indicators can be found in~\tref{table:genera_4}.
~\fref{fig:rader1} provides visualization of our generalization tests.

\begin{figure*}[t] 
    \centering
    \includegraphics[width=.99\linewidth]{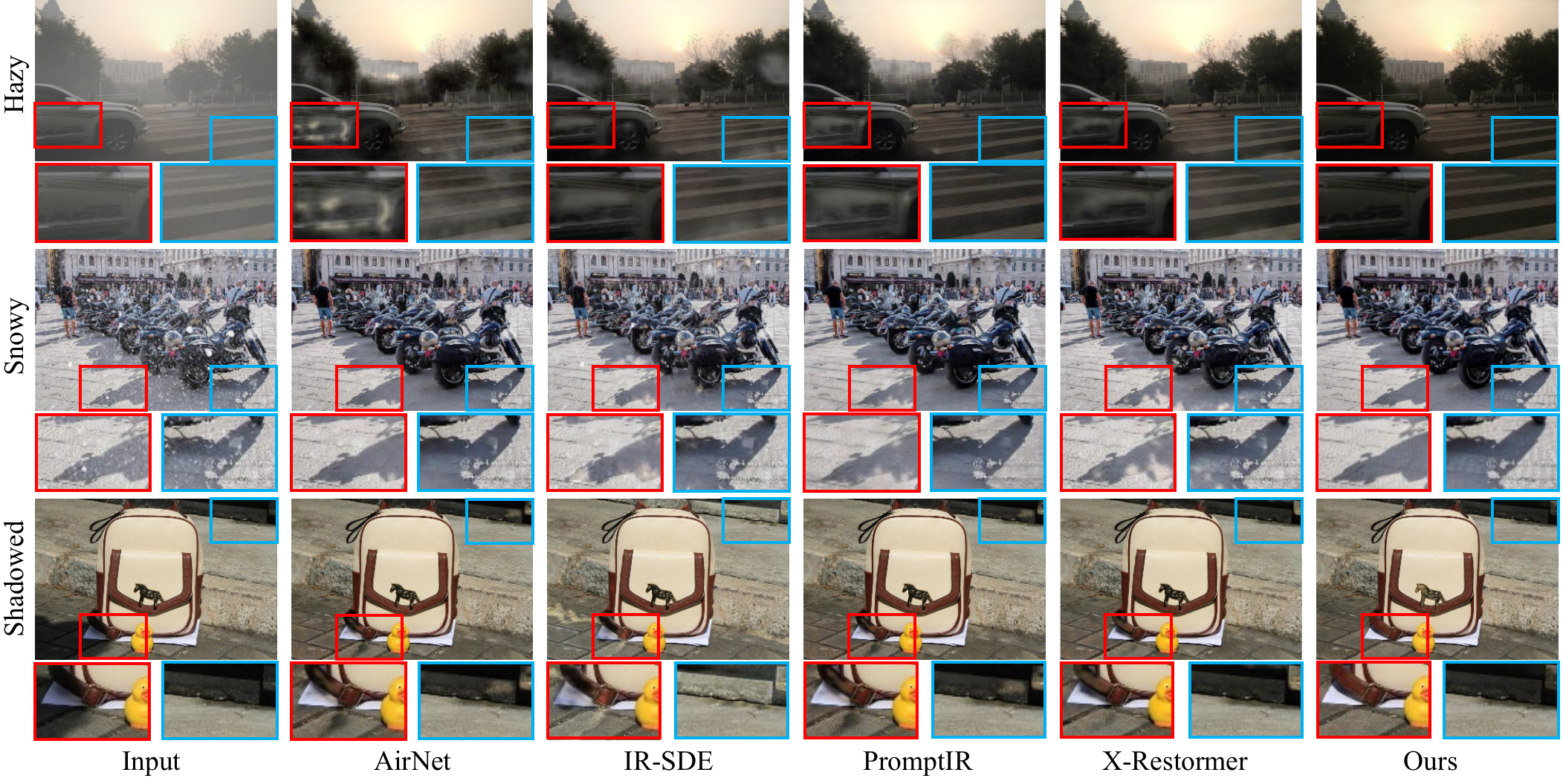}
    \vspace*{-1mm}
    \caption{
    Visualization comparisons of our method with previous approaches for \textcolor{blue! 100}{Setting B} in \tref{table:deg10}. Our method can obtain cleaner images while preserving more background information when dealing with complex and multiple types of degradation. 
    More visualization details in supplementary material.}
    \vspace*{-1em}
    \label{fig:vision}
\end{figure*}

\noindent
\textbf{Comparison in} \textcolor{blue! 100}{Setting B}.
We evaluated the model’s multi-task unified IR capability on a mixture of 10 degradation types of data. The detailed indicator comparison is shown in \tref{table:deg10}. 
Our method achieves the best results in both fidelity and perceptual indicators. At the same time, our method can be extended to the existing image restoration network and further improves the restoration performance.
Our restoration effects and indicator radar charts for ten degradation types can be found in \fref{fig:rader1}.
\fref{fig:vision} shows the restoration effect comparison with the SOTA methods. Our method can effectively restore cleaner images from ten types of degraded images.

\begin{figure}[tbp]
\begin{minipage}[t]{0.45\columnwidth}
\vspace{0cm}
\setlength{\abovecaptionskip}{0.1cm}
\setlength{\belowcaptionskip}{0.1cm}
\begin{center}
\captionsetup{type=table}
\caption{Ablation study of main work. }
\label{table:ablation main}
\setlength{\tabcolsep}{2pt}
\scalebox{0.6}{
    \begin{tabular}{c|cc|cc|ccc}
    \toprule
    \multirow{3}[6]{*}{\textbf{Method}} & \multicolumn{4}{c|}{\textcolor{blue! 100}{Setting A}} & \multicolumn{3}{c}{\textcolor{blue! 100}{Setting B}} \\
\cmidrule{2-8}          & \multicolumn{2}{c|}{Restoration} & \multicolumn{2}{c|}{Generalization} & \multicolumn{3}{c}{Average performance} \\
\cmidrule{2-8}          & PSNR↑ & SSIM↑ & PSNR↑ & SSIM↑ & PSNR↑ & SSIM↑ & LPIPS↓ \\
    \midrule
    w/o TMGP & 28.95 & 0.895 & 15.29  & 0.535 & 24.56 & 0.782 & 0.464 \\
    w/o GEF & 21.25 & 0.624 & 14.47  & 0.565 & 17.76 & 0.564 & 0.529 \\
    w/o TIT & -     & -     & -     & -     & 27.01 & 0.743 & 0.394 \\
    w/o MoE & -     & -     & -     & -     & 27.64 & 0.842 & 0.188 \\
    w/o $\mathcal{L}_{reg}$ & 28.65 & 0.844 & 16.10  & 0.611 & 27.41 & 0.794 & 0.236 \\
    w/o $\mathcal{L}_{orthog}$ & 28.79 & 0.909 & 19.88  & 0.742 & 27.25 & 0.812 & 0.252 \\
    \rowcolor{blue! 8} Ours  & \textbf{29.08} & \textbf{0.933} & \textbf{20.74 } & \textbf{0.782} & \textbf{28.13} & \textbf{0.862} & \textbf{0.123} \\
\bottomrule
\end{tabular}}
\end{center}
\end{minipage}
\hfill
\begin{minipage}[t]{0.52\columnwidth}
\vspace{0mm}
\setlength{\abovecaptionskip}{0.2cm}
\setlength{\belowcaptionskip}{0.2cm}
\centering
    \includegraphics[width=.93\linewidth]{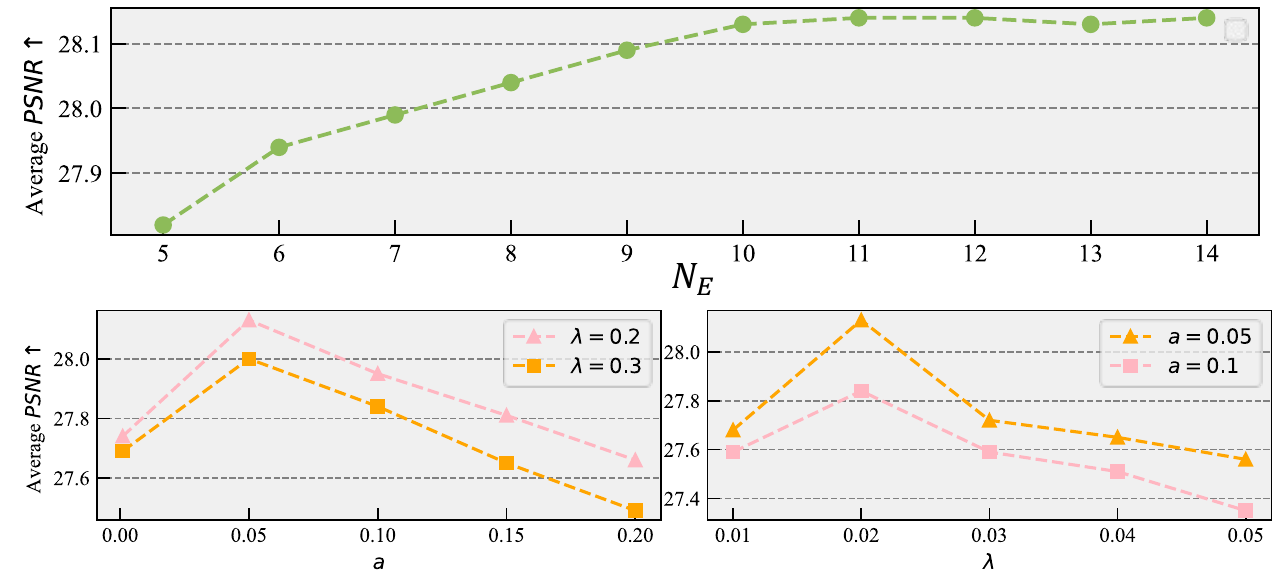}
    \vspace{-1.5mm}
    \caption{Ablation of loss and hyperparameters.}  
    \label{ablation_hy}
\end{minipage}
\vspace{-2em}
\end{figure}

\noindent
\textbf{Ablation study.}
\label{Ablation}
For single-task generalization in \textcolor{blue! 100}{Setting A}, we take dehazing as an example to conduct an ablation study on Timestep Matched Generative Pre-training (TMGP), Generalization Enhanced Fine-tuning (GEF), and loss functions.
For multi-task restoration in \textcolor{blue! 100}{Setting B}, we added Time-sequential Incremental Training (TIT) and MoE Adapters. 
As shown in \tref{table:ablation main}, TMGP and GEF play a key role in improving the generalization of the model. The loss function can effectively enhance the model performance, which is consistent with our analysis in \sref{Analysis}. The incremental training paradigm can effectively improve multi-task performance, and MoE can further enhance this.
In addition, we conduct an ablation study on the hyperparameters $\lambda$ and $a$ of the loss function and $N_E$ of MoE in \textcolor{blue! 100}{Setting B}. As shown in \fref{ablation_hy}, when experts increases to a certain value, its sensitivity decreases, and we take the value with the best PSNR performance for $\lambda$ and $a$. 
More details on the implementation of the ablation experiments are provided in the \textbf{supplementary material}.

\vspace*{-4mm}

\section{Related Work}
\vspace{-1.1em}
\textbf{General image restoration.}
Advancements in image restoration (IR) primarily utilize neural networks for pixel-wise learning from paired images~\cite{task9711014,Xiao9710011Generalization,Xiao2023RandomST,Xiao_2024_CVPR}, and advanced network architectures are increasingly applied to multiple restoration tasks like deraining \cite{Fu8099669}, desnowing \cite{Snow01_2021}, dehazing \cite{Haze01_2022} and low-light enhancement~\cite{Yang2021SparseGR}.
While they eliminate the need for specific task priors, addressing multiple degradation types requires distinct training procedures. 
Moreover, the generalization of single-task IR poses a significant challenge~\cite{Xiao9710011Generalization}.
To tackle the challenge of a single model handling multiple degradations, several unified IR methods have been successively proposed~\cite{Zhu2023LearningWA,yang2023language,qin2024restoremasksleveragingmask,luo2024controlling}. 
These methods train on mixed paired images from diverse image restoration tasks to minimize the gap between different degradation types. Despite the remarkable efficacy, they are still constrained by limited degradation types and specific task dataset constraints~\cite{luo2024controlling}. 
Therefore, there is an urgent need for a simple yet effective approach to enhance the single-task generalization capability of general IR networks and improve the stability in addressing multi-task unified IR.

\noindent
\textbf{Image restoration with diffusion.}
Diffusion models are prominent solution in image generation, modeling the forward diffusion akin to a Wiener process and learning the denoising reverse process~\cite{sohl2015deep, ho2020denoising, song2019generative, song2020improved, song2021denoising, song2021maximum, song2021score, rombach2022high,rissanen2022generative}.
Due to their strong generative capabilities, researchers leverage diffusion to design generative-based image restoration methods~\cite{AutoDIR,DiffIR,luo2023image}. 
\citet{lugmayr2022repaint} uses a pre-trained generative model directly to generate clean images, while \citet{DiffUIR} retrains a diffusion model with specific mapping rules to generate clean images from low-quality images. Although enhancing perceptual restoration performance, they introduce architectural complexity and low generation efficiency from image generation~\cite{DreamClean}.
Some recent some studies integrate the diffusion mechanism into the training loss of reconstruction-based IR networks to enhance the generalization and adaptation capabilities~\cite{tan2024diffloss}.
Nevertheless, more exploration is required to understand the diffusion mechanism's impact on reconstruction-based IR networks and optimize its generative potential for improving general IR networks' performance.
To address the above challenges, we comprehensively analyze the diffusion mechanism' impact on reconstruction-based IR from multiple perspectives.
By combining the diffusion generative objective with the IR reconstruction objective, we propose a new diffusion training enhanced IR framework for elevating general IR models.
\vspace*{-4mm}
\section{Conclusion}
\vspace*{-3mm}

This paper revisits the application of the generation mechanism of diffusion models in general restoration networks, links the generation objective with the reconstruction objective to help the restoration network learn the prior information in the diffusion latent space. 
We propose the diffusion-based generative pre-training and adaptive fine-tuning based on degradation-specific matching time step $t$ to enhance performance and generalization in single-task restoration. 
For multi-task restoration, we propose an incremental training strategy guided by degradation-specific matching time steps and introduce MoE adapters with time-based prompts to further enhance the unified restoration performance.
Extensive experiments demonstrate the effectiveness of our approach, giving a new perspective for the application of diffusion models in image restoration.

\newpage

\appendix

\section{Appendix}

In the Technical Appendices, we offer the details omitted from the main text, summarizing this supplemental content into these sections: (1) Sec. \ref{Supplement to The Text}, Supplement to The Text; (2) Sec. \ref{Future Work and Limitation.}, Future Work and Limitation; (3) Sec. \ref{Societal Impact.}, Broader Impacts.

Due to file size limitations, more visualizations and detailed discussions are provided in the \textbf{supplementary material}.

\section{Supplement to The Text}
\label{Supplement to The Text}
\begin{figure*}[ht] 
    \centering
    \includegraphics[width=1\linewidth]{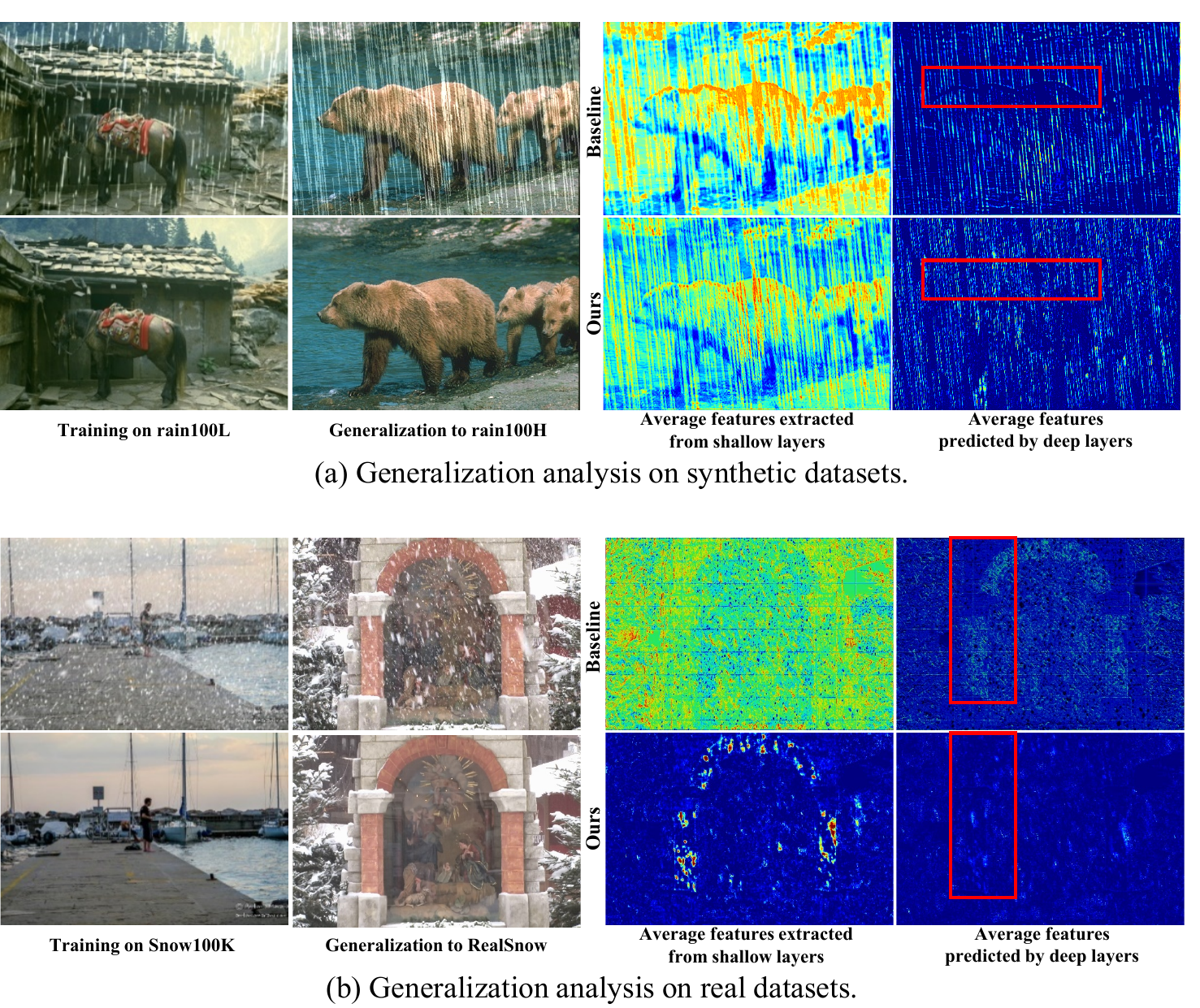}
    \caption{
    \textbf{Comparison of shallow and deep features in generalization evaluations on synthetic and real datasets.} The IR network extracts image information in shallow layer and outputs degraded residuals in deep layer. Our method reduces the damage to the background pixels of the image by learning the diffusion's generative representation.
    (a)  Generalized to synthetic dataset, degraded residual output by the baseline can remove rain streaks but also causes the loss of background. Our method predicts the degradation information more accurately.
    (b) Generalized to real dataset, baseline can't handle unseen complex real data, our method maintains restoration ability.
    }
    \label{fig:layer_feature_real}
\end{figure*}

\tref{table:dataset} and \fref{fig:layer_feature_real} supplement the paper’s settings for the multi-task unified image restoration task and the sensitivity of different layers of the image restoration network to diffusion generation training.

\begin{table*}[ht]
\caption{\underline{\textbf{Multi-task unified IR}} (\textcolor{blue! 100}{Setting B}): 10 restoration tasks' data sources and the matching $t^{mat}$ for each degradation type.}
\label{table:dataset}
\centering
\resizebox{1\linewidth}{!}{
\begin{tabular}{ccccccccccc}
\toprule
Tasks & \makecell[c]{Noisy}   & \makecell[c]{Rainy}   & \makecell[c]{JPEG}    & \makecell[c]{Snowy}  & \makecell[c]{Inpainting}   & \makecell[c]{Raindrop}   & \makecell[c]{Shadowed}    & \makecell[c]{Low-light}  & \makecell[c]{Hazy}   & \makecell[c]{Blurry}    \\ \midrule
$t^{mat}$  &4     & 8     & 12    & 15    & 19    & 22    & 27    & 38    & 47    & 50\\
Source & DIV2K~\citep{agustsson2017ntire}, Flickr2K~\citep{timofte2017ntire}, CBSD68~\citep{martin2001database}& Rain100H~\citep{yang2017deep} & DIV2K~\citep{agustsson2017ntire}, Flickr2K~\citep{timofte2017ntire},LIVE1~\citep{sheikh2005live} & Snow100K~\citep{Snow100K2018} & RePaint~\citep{lugmayr2022repaint} & RainDrop~\citep{Qian2017AttentiveGA} & SRD~\citep{qu2017deshadownet} & LOL~\citep{wei2018deep} & RESIDE~\cite{OTS2019} & GoPro~\citep{nah2017deep} \\
\bottomrule
\end{tabular}
}
\vspace*{-2.0mm}
\end{table*}

\begin{figure*}[ht] 
    \centering
    \includegraphics[width=1\linewidth]{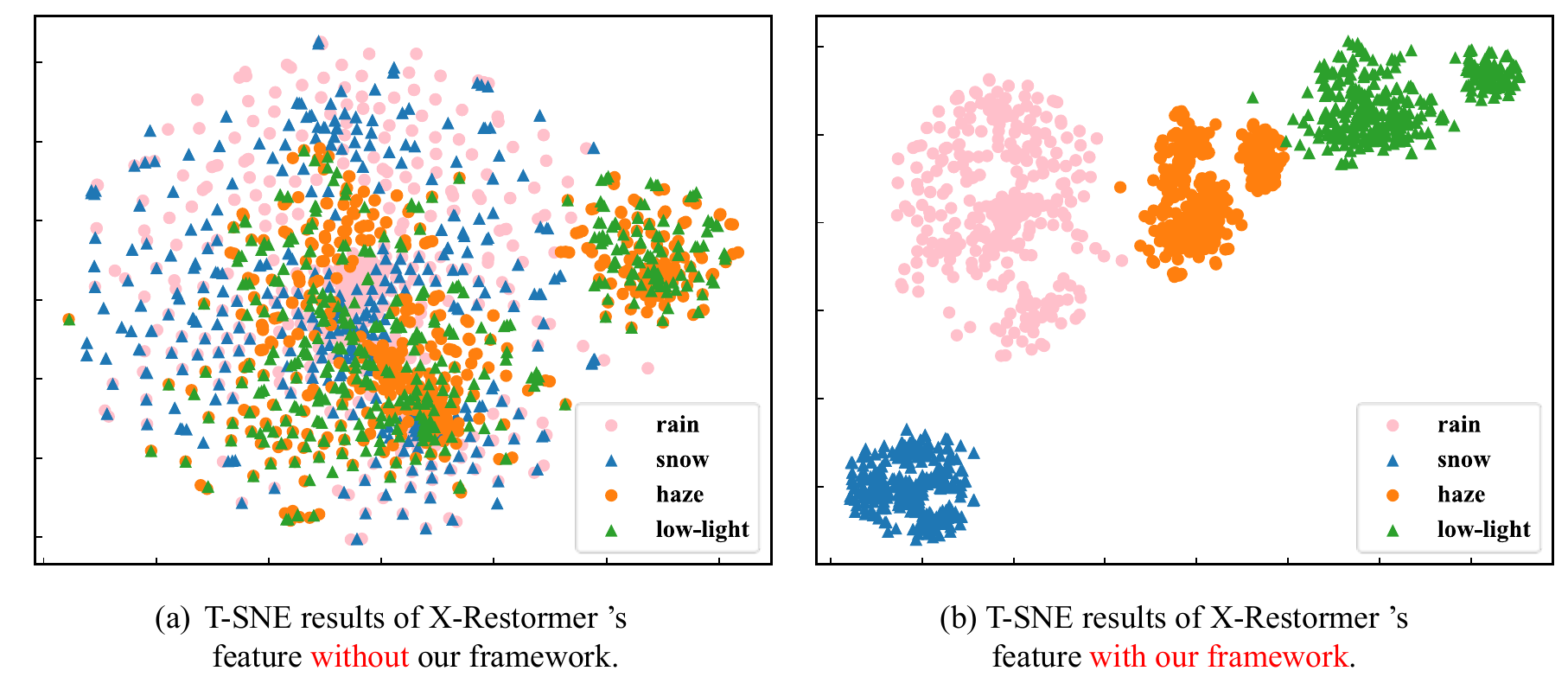}
    \caption{
    \textbf{T-SNE results} of general IR Network X-Restormer’s feature (left) and feature of X-Restormer~\cite{X-Restormer} using our framework (right).    }
    \label{fig:tsne1}
\end{figure*}

We perform T-SNE analysis on multiple degradation types in \underline{\textbf{Multi-task unified IR}} (\textcolor{blue! 100}{Setting B}). It can be found that general image restoration network X-Restormer~\cite{X-Restormer} that has not been trained by our framework cannot distinguish different heterogeneous degradation types well, which leads to the deterioration of the integrated image restoration effect. By matching the optimal time step by our generative pre-training, general restoration networks that perform task-specific fine-tuning can clearly distinguish different degradation types, which verifies our view that optimal time step helps adaptively identify different degradation types, thereby significantly enhancing the multi-task unified image restoration effect.
\section{Future Work and Limitation}
\label{Future Work and Limitation}
While our approach can be easily transferred to various general restoration networks, the diffusion-based training method still incurs additional time consumption in practical applications. Our future work will focus on enhancing this training strategy, devising new unified training methods to improve model performance while reducing training duration.

\section{Broader Impacts}
\label{Societal Impact}
In today's world, image capture systems inevitably face a range of degradation issues, stemming from inherent noise in imaging devices, capture instabilities, to unpredictable weather conditions. Consequently, image restoration holds significant research and practical value. Our proposed Diffusion Training Enhanced IR Framework enables image restoration networks to refine degraded images more effectively. However, from a societal perspective, negative consequences may also arise. For instance, the actual texture discrepancies introduced by image restoration techniques could impact fair judgments in medical and criminal contexts. In such scenarios, it becomes imperative to integrate expert knowledge for making informed decisions.

\newpage
\section{supplementary materials}
\label{supplementary materials}

In the supplementary materials, we offer the details omitted from the main text, summarizing this supplemental content into these sections: 

(1) Sec. \ref{Algorithm Analysis}, Exposition of Algorithm, including more insightful analysis and algorithm implementation details as well as theoretical derivations; 

(2) Sec. \ref{Details}, More Details, including experimental and dataset setup, training details, and other supplements to the main text; 

(3) Sec. \ref{Further Study Investigations}, Further Study Investigations, including model efficiency and complexity analysis, more supplementary ablation experiments, quantitative and qualitative analysis, and more visualization results.; 

(4) Sec. \ref{Future Work and Limitation.}, Future Work and Limitation; 

(5) Sec. \ref{Societal Impact.}, Broader Impacts.

\begin{figure}[ht]
    \centering
    \includegraphics[width=1.\linewidth]{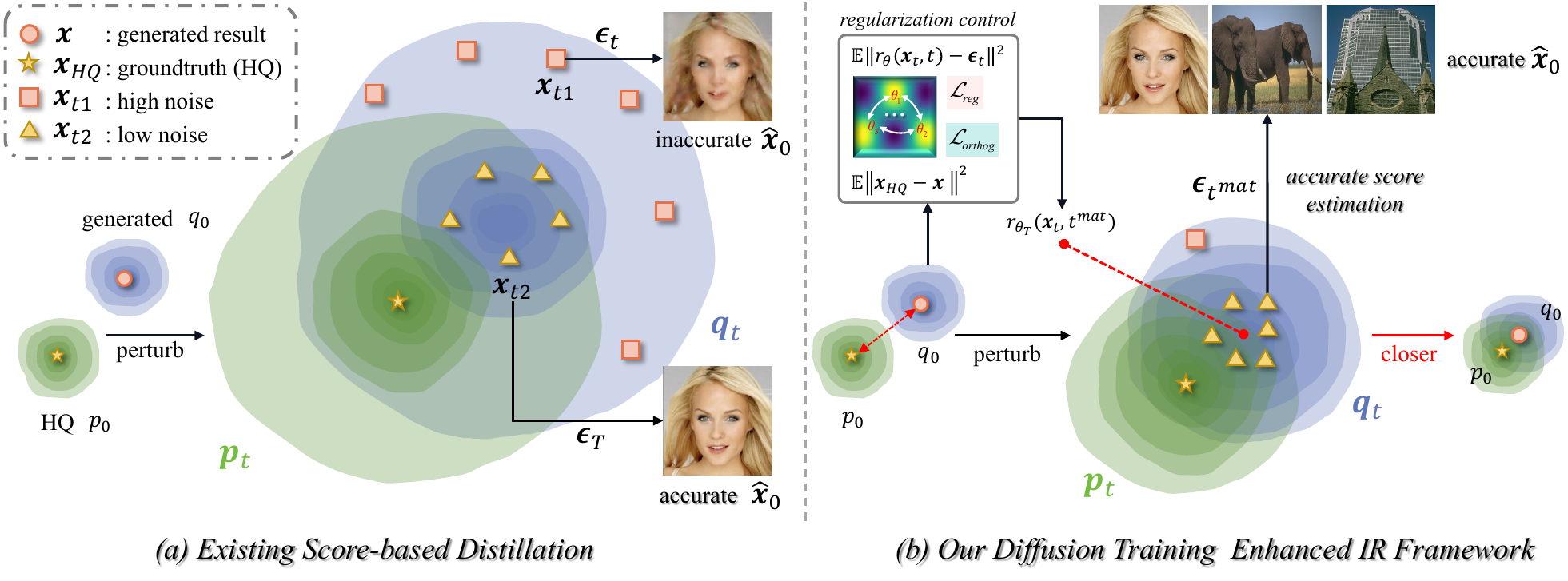}
    \caption{Comparison with existing diffusion-based approaches from a score-based perspective. (a) Existing diffusion-based approaches use score-based distillation to use a wide range of perturbations, resulting in large noise that makes the generated results $\bm x$ far away from the GT. This leads to inaccurate score estimation (described by low-quality pseudo GT $\hat{\bm x}_0$) and hinders image reconstruction. 
     (b) Our IR framework based on diffusion training enhancement adaptively adjusts the image reconstruction objective and score matching objective, uses a series of regularization strategies combined with matching time steps to control the scale of noise perturbation, and achieves a tighter score distribution by aligning the diffusion generation score estimate to the image reconstruction, thereby enabling the image restoration network $r_{\theta_{\mathcal{T}}}(x_t,t^{mat})$ to achieve accurate image reconstruction $\hat{\bm x}_0$.  }
    \label{fig:score_vision}
\end{figure}

\section{Exposition of Algorithm}
\label{Algorithm Analysis}

\subsection{Insights From the Perspective of Score Matching}
\label{Insights}
To present our core insights more clearly, in this section we analyze in detail the differences between our \textit{Diffusion Training Enhanced IR Framework} and previous diffusion-based methods, and reveal the underlying principles of this new framework from a score-based perspective.

As shown in \fref{fig:score_vision}, our IR framework based on diffusion training enhancement adaptively adjusts the image reconstruction objective and score matching objective, uses a series of regularization strategies combined with matching time steps to control the scale of noise perturbation, and achieves a tighter score distribution by aligning the diffusion generation score estimate to the image reconstruction, thereby enabling the image restoration network $r_{\theta_{\mathcal{T}}}(x_t,t^{mat})$ to achieve accurate image reconstruction $\hat{\bm x}_0$.
This strategy enables our framework to achieve highly competitive performance on image restoration tasks with 10 degradation types.

\subsection{Algorithm Implementation Details}
\label{Algorithm}
The image restoration algorithm framework proposed in this paper mainly consists of diffusion generative pre-training, task-specific image restoration fine-tuning, and multi-task incremental unified learning. The pseudo code of diffusion generative pre-training is shown in Algorithm \ref{alg:DDPM}, matching the optimal time step reverse diffusion sampling is shown in Algorithm \ref{alg:sampling}, generalization enhanced task-specific image restoration fine-tuning is shown in Algorithm \ref{alg:Fine-tuning}, and the effective multi-task incremental unified learning method is shown in Algorithm \ref{alg:Multi-task}. Our overall framework of image restoration enhanced by diffusion training is shown in Algorithm \ref{alg:Framework}.

\begin{algorithm}[ht]
  \caption{Diffusion Generative Pre-training} \label{alg:DDPM} 
  \begin{algorithmic}[1]
    \Require $\mathbf x_0$, $\mathbf x_T$
    \Require a image restoration network $r_\theta$
      \State $\mathbf x_0 \sim q(\mathbf x_0)$
      \State $\mathbf x_T \sim \mathcal{N}(\mathbf{0}, \mathbf I)$      
      \State $\mathbf \epsilon\sim\mathcal{N}(\mathbf {0},\mathbf I)$
      \State $t \sim \mathrm{Uniform}(\{1, \dotsc, T\})$
    \Repeat
      \State Take gradient descent step on
      \Statex $\qquad \nabla_\theta \left\| \mathbf \epsilon - r_\theta(\sqrt{\bar\alpha_t} \mathbf x_0 + \sqrt{1-\bar\alpha_t}\mathbf \epsilon, t) \right\|^2$
    \Until{converged}
    \State \textbf{return} $r_{\theta_{pre}}$
  \end{algorithmic}
\end{algorithm}

\begin{algorithm}[ht]
  \caption{Timestep Matched Sampling} \label{alg:sampling}
  \begin{algorithmic}[1]
    \Require $\mathbf x_0$, $\mathbf x_t$, $\mathbf x_T$, $\mathbf x$, $\mathbf y$
    \Require a image restoration network $r_\theta$
    \Require a generative pre-trained image restoration network $r_{\theta_{pre}}$
    \State $\mathbf x_0 \sim q(\mathbf x_0)$
    \State $\mathbf x_T \sim \mathcal{N}(\mathbf{0}, \mathbf I)$
    \State $\mathbf y \sim p(\mathbf y)$
    \State $\mathbf x \sim { r_\theta}^{-1}(\mathbf y)$
    \State $t \sim \mathrm{Uniform}(\{1, \dotsc, T\})$
    \For{$t=T, \dotsc, 1$}
      \State $\mathbf z \sim \mathcal{N}(\mathbf{0}, \mathbf I)$ \textbf{If} $t > 1$, else $\mathbf z = \mathbf{0}$
      \State $\mathbf x_{t-1} = \frac{1}{\sqrt{\alpha_t}}\left( \mathbf x_t - \frac{1-\alpha_t}{\sqrt{1-\bar\alpha_t}} \mathbf \epsilon_\theta(\mathbf x_t, t) \right) + \sigma_t \mathbf z$   
      \State $t^{mat} \leftarrow t$ \textbf{If} min $ \left\| (\mathbf x_{t} - \mathbf x_{t-1}) - (\mathbf y - \mathbf x)  \right\|^2$
    \EndFor  
    \State \textbf{return} $\mathbf x_0$, $t^{mat}$ 
  \end{algorithmic}
\end{algorithm}

\begin{algorithm}[ht]
  \caption{Generalization Enhanced Fine-tuning} \label{alg:Fine-tuning}
  \begin{algorithmic}[1]
    \Require $\mathbf x_0$,  $\mathbf x$, $\mathbf y$
    \Require a image restoration network $r_\theta$ and a generative pre-trained image restoration network $r_{\theta_{pre}}$
    \Require training diffusion generation phase task $T_0$, fine-tuning phase task $T_{1}$
    \Require generation training loss $\mathcal{L}_g$, fine-tuning training loss $\mathcal{L}_{fine}$
    \State $\mathbf x_0 \sim q(\mathbf x_0)$
    \State $\mathbf y \sim p(\mathbf y)$
    \State $\mathbf x \sim { r_\theta}^{-1}(\mathbf y)$
    \For{training pairs $\{\mathbf x_i,\mathbf y_i\}_1^n$}
      \State Training Task $T_0$ using generation loss $\mathcal{L}_g$
          \If{$i \in {B_g}$ in Eq.\ref{s}}
          \For{Parameter ${\theta_{pre}}$}
          \State ${\theta_{pre}} \leftarrow {\theta_{pre}} + \epsilon\nabla \mathcal{L}_g$
          \EndFor  
          \EndIf
      \State Training Task $T_1$ using fine-tuning loss $\mathcal{L}_{fine}$
          \If{$i \in {B_r}$ in Eq.\ref{s}}
          \For{Parameter ${\theta_{pre}}$}
          \State ${\theta_{fine}} \leftarrow {\theta_{pre}} + \epsilon\nabla \mathcal{L}_{fine}$
          \EndFor  
          \EndIf
    \EndFor  
    \State \textbf{return} A task-specific generalization enhanced  fine-tuned network $r_{\theta_{fine}}$ 
  \end{algorithmic}
\end{algorithm}

\begin{algorithm}[ht]
  \caption{Effective Multi-task Unified Learning} 
  \label{alg:Multi-task}
  \begin{algorithmic}[1]
    \Require $\mathcal{T}={\{t_i^{mat}\}}_{i=1}^n$, $\mathbf x_0$, $\mathbf x$, $\mathbf y$
    \Require a image restoration network $r_\theta$, incremental training image restoration network $\{r_{\theta_i}\}_1^n$
    \Require a generative pre-trained image restoration network 
    $r_{\theta_{pre}}$
    \Require a task-specific image restoration fine-tuning network $r_{\theta_{fine}}$
    \State $\mathbf x_0 \sim q(\mathbf x_0)$
    \State $\mathbf y \sim p(\mathbf y)$
    \State $\mathbf x \sim { r_\theta}^{-1}(\mathbf y)$
    \State $i \sim \mathrm{Uniform}(\{1, \dotsc, n\})$
    \For{$i=n, \dotsc, 1$}
    \For{training pairs $\{\mathbf x_j,\mathbf y_j\}_1^{n_t}$ with $t_i^{mat}$}
          \If{$j \in {B_g}$ in Eq.\ref{s}}
          \For{Parameter ${\theta_{i}}$}
          \State ${\theta_{i}} \leftarrow {\theta_{i-1}} + \epsilon (\nabla_{\theta_{pre}} \mathcal{L}_{fine}+\nabla_{\theta_{i-1}} \mathcal{L}_g)$
          \EndFor  
          \EndIf
          \If{$j \in {B_r}$ in Eq.\ref{s}}
          \For{Parameter ${\theta_{i}}$}
          \State ${\theta_{i}} \leftarrow {\theta_{i-1}} + \epsilon(\nabla_{\theta_{pre}} \mathcal{L}_{fine} + \nabla_{\theta_{i-1}} \mathcal{L}_{fine})$
          \EndFor  
          \EndIf
    \EndFor  
    \EndFor 
    \State \textbf{return} Multi-task Unified IR network $r_{\theta_{\mathcal{T}}}$
  \end{algorithmic}
\end{algorithm}

\begin{algorithm}[ht]
  \caption{Our Diffusion Training Enhanced IR Framework} 
  \label{alg:Framework}
  \begin{algorithmic}[1]
    \Require $\mathcal{T}={\{t_i^{mat}\}}_{i=1}^n$, $\mathbf x_0$, $\mathbf x_t$, $\mathbf x_T$, $\mathbf x$, $\mathbf y$
    \Require a image restoration network $r_\theta$, incremental training image restoration network $\{r_{\theta_i}\}_1^n$
    \Require a generative pre-trained image restoration network 
    $r_{\theta_{pre}}$
    \Require a task-specific image restoration fine-tuning network $r_{\theta_{fine}}$
    \Require generation training loss $\mathcal{L}_g$, fine-tuning training loss $\mathcal{L}_{fine}$
    \State $\mathbf x_0 \sim q(\mathbf x_0)$
    \State $\mathbf x_T \sim \mathcal{N}(\mathbf{0}, \mathbf I)$
    \State $\mathbf y \sim p(\mathbf y)$
    \State $\mathbf x \sim { r_\theta}^{-1}(\mathbf y)$
    \State $t \sim \mathrm{Uniform}(\{1, \dotsc, T\})$
    \State $i \sim \mathrm{Uniform}(\{1, \dotsc, n\})$
    \Repeat
      \State Take gradient descent step on
      \Statex $\qquad \nabla_\theta \left\| \mathbf \epsilon - r_\theta(\sqrt{\bar\alpha_t} \mathbf x_0 + \sqrt{1-\bar\alpha_t}\mathbf \epsilon, t) \right\|^2$
    \Until{converged}
    \For{$t=T, \dotsc, 1$}
      \State $\mathbf z \sim \mathcal{N}(\mathbf{0}, \mathbf I)$ \textbf{If} $t > 1$, else $\mathbf z = \mathbf{0}$
      \State $\mathbf x_{t-1} = \frac{1}{\sqrt{\alpha_t}}\left( \mathbf x_t - \frac{1-\alpha_t}{\sqrt{1-\bar\alpha_t}} \mathbf \epsilon_\theta(\mathbf x_t, t) \right) + \sigma_t \mathbf z$   
      \State $t^{mat} \leftarrow t$ \textbf{If} min $ \left\| (\mathbf x_{t} - \mathbf x_{t-1}) - (\mathbf y - \mathbf x)  \right\|^2$
    \EndFor  
    \State \textbf{return} $\mathbf x_0$, $t^{mat}$, $r_{\theta_{fine}}$
    
    \For{$i=n, \dotsc, 1$}
    
    \For{training pairs $\{\mathbf x_j,\mathbf y_j\}_1^{n_t}$ with $t_i^{mat}$}
          \If{$j \in {B_g}$ in Eq.\ref{s}}
          \For{Parameter ${\theta_{i}}$}
          \If{$i>1$}
          \State ${\theta_{i}} \leftarrow {\theta_{i-1}} + \epsilon (\nabla_{\theta_{pre}} \mathcal{L}_{fine}+\nabla_{\theta_{i-1}} \mathcal{L}_g)$
          \Else 
          \State ${\theta_{i}} \leftarrow {\theta_{pre}} + \epsilon\nabla \mathcal{L}_g$
          \EndIf
          \EndFor  
          \EndIf
          \If{$j \in {B_r}$ in Eq.\ref{s}}
          \For{Parameter ${\theta_{i}}$}
          \If{$i>1$}
          \State ${\theta_{i}} \leftarrow {\theta_{i-1}} + \epsilon(\nabla_{\theta_{pre}} \mathcal{L}_{fine} + \nabla_{\theta_{i-1}} \mathcal{L}_{fine})$
        \Else 
        \State ${\theta_{i}} \leftarrow {\theta_{pre}} + \epsilon\nabla \mathcal{L}_{fine}$
        \EndIf
          \EndFor  
          \EndIf
    \EndFor  
    \EndFor 
    \State ${\theta_{\mathcal{T}}} \leftarrow {\theta_{n}}  $
    \State \textbf{return} Multi-task Unified IR network $r_{\theta_{\mathcal{T}}}$

  \end{algorithmic}
\end{algorithm}

Similar to DDPM~\cite{ho2020denoising}, our optimization objective of diffusion generative pre-training is:
\begin{equation}
    \mathcal{L}_{\gamma}(\theta) \coloneqq \sum_{t=1}^T \gamma_t \, \mathbb{E} \Big[ \big\lVert \tilde{r}_\theta({x}_{t}, x_0, t) - \epsilon_{t} \bigr\rVert\Big],
    \label{eq:noise_objective}
\end{equation}
where $\gamma_1, \ldots, \gamma_T$ are loss weights.
By applying diffusion and reverse diffusion process for degradation image with different time $t$, we can find $t_i^{mat}$ that best matches reversed image with clean image, which helps us distinguish different degradation types and match them to the latent space of diffusion.

To address the gap between the diffusion generation objective and the image restoration reconstruction objective during task-specific image restoration fine-tuning with enhanced generalization, we propose parameter importance regularization and gradient orthogonality techniques. 
The specific implementation details of these regularization strategies can be found in Algorithm \ref{alg:Framework}
Inspired by continual learning \cite{EWC2017}, we utilize a parameter regularization strategy with second-order parameter importance modification schemes.
We first calculate the parameter importance of the IR network obtained by generative pre-training, then update the network parameters according to the importance weights in fine-tuning. 
Specifically, denoting pre-training as task 0, fine-tuning as task 1, parameter importance weights $\Omega_{\theta_{k}}$  are calculated by accumulating gradients:
\begin{equation}
    \begin{split}        \Omega_{\theta_{k}}=f(x;\theta_{r}^{1})-f(x;\theta_{k}^{0}),
    \end{split}
    \label{importance1}
\end{equation}
where $f(\cdot)$ represents the mapping function of IR network, $\theta_{k}$ denotes $k$th parameter of the IR network, and $\delta\theta_{k}^{1}=\delta\theta_{k}^{0}+\delta\theta_{k}$. $\delta$ denotes the parameter change magnitude, and indicates the input various degradation data. In particular, the above equation can be written as:
\vspace*{-0.5em}
\begin{equation}
    \begin{split}        \Omega_{\theta_{k}}=\nabla_{\theta_{k}}\mathcal{L}\,|\delta\theta_{k}|+\frac{1}{2}\cdot\nabla_{\theta_{k}}^{2}\mathcal{L}\,|\delta\theta_{k}|^{2}+O(|\delta\theta_{k}|^{3}),
    \end{split}
    \label{importance2}
\end{equation}
where $\mathcal{L}$ is the conventional loss of baseline method. 
We retain the first two terms as second-order importance and obtain the parameter regularization loss that maintains the generalization ability of generative pre-training in restoration fine-tuning as follows:
\begin{equation}
    \begin{split}        
    {\mathcal{L}_{reg}}=\lambda\sum_{k=1}^{m}\left[\nabla_{\theta_{k}}{\mathcal{L}}\,|\delta\theta_{k}|+{\frac{1}{2}}\cdot\nabla_{\theta_{k}}^{2}\,{\mathcal{L}}\,|\delta\theta_{k}|^{2}\right],
    \end{split}
    \label{importance}
\end{equation}
where $\lambda$ is balance coefficient, $m$ is number of parameters.

During mixed generative task's restoration fine-tuning process, we incorporate gradient orthogonality to further ensure training efficiency.
Given input-gt pairs $\{{x}_i,y_i\}_{i=0}^B$, generative training with $i \in {B_g}$ and reconstruction training with $i \in {B_r}$. Let $\mathbf{g}_i=\nabla_\theta\mathcal{L}=\partial \mathcal{L}/\partial \theta$ be the gradient vector during fine-tuning. 
We design a unified gradient orthogonal loss $\mathcal{L}_{orthog}$ to constrain the gradient direction of generation and reconstruction during fine-tuning as follows:
\begin{equation}
    \begin{split}        
s = \sum\limits_{i \in {B_g},j \in {B_r}} {\left\langle {{\mathbf{g}_i},{\mathbf{g}_j}} \right\rangle /\sum\limits_{i \in {B_g},j \in {B_r}} 1 },
    \end{split}
    \label{s}
\end{equation}
\begin{equation}
    \begin{split}        
d = \left( {\sum\limits_{i,k \in {B_g}} {\left\langle {{\mathbf{g}_i},{\mathbf{g}_k}} \right\rangle }  + \sum\limits_{l,j \in {B_r}} {\left\langle {{\mathbf{g}_l},{\mathbf{g}_j}} \right\rangle } } \right)/\sum\limits_{i,k,l,j \in {B}} 1 ,
    \end{split}
    \label{d}
\end{equation}
\begin{equation}
    \begin{split}        
{\mathcal{L}_{orthog}} = \left( {1 - s} \right) + \left| d \right|,
    \end{split}
    \label{zjloss}
\end{equation}
 where $\left\langle { \cdot , \cdot } \right\rangle$ is the cosine similarity operator applied on two
 vectors, $\left|  \cdot  \right|$ is the absolute value operator, and $B$ is
 mini-batch size. Note that the cosine similarity operator 
 used in Eq. \ref{s} and Eq. \ref{d} involves normalization of features (projection to a unit hyper-sphere) as follows:
\begin{equation}
    \begin{split}        
\left\langle {{\mathbf{g}_i},{\mathbf{g}_j}} \right\rangle  = \frac{{{\mathbf{g}_i} \cdot {\mathbf{g}_j}}}{{{{\left\| {{\mathbf{g}_i}} \right\|}_2} \cdot {{\left\| {{\mathbf{g}_j}} \right\|}_2}}},
    \end{split}
    \label{cos}
\end{equation}
where $\left\|  \cdot  \right\|_2$ refers to the $\ell_2$ norm operator.

Since different layers of the network have different requirements for generation capabilities (for detailed analysis, please refer to the main text), we perform gradient updates only in the shallow layers which sensitive to generative objectives but less affect reconstructive objectives and thus constrain the gap, applying weight decay to the parameters with matching time step $t^{mat}$: 
\begin{equation}
    \begin{split}        
w_{decay}(t) = e^{-at} ,
    \end{split}
    \label{decay}
\end{equation}
where a is the decay rate, which is set to 0.05.
During the reconstructive training in the fine-tuning process, we also adjust the balancing coefficient $\lambda$  for the regularization loss $\mathcal{L}_{reg}$ in Eq. \ref{importance} based on the layer-wise decay of the network when applying parameter importance regularization.

Parameter importance guided updates in \eref{importance2} avoid the model forgetting previously generative knowledge, gradient orthogonality in \eref{zjloss} enhances generalization capabilities through mixed generation training. 
Total optimization objective for fine-tuning is defined as follows:
\begin{equation}    \mathcal{L}_{fine}=\mathcal{L}_{content}+\mathcal{L}_{reg}+\mathcal{L}_{orthog},
    \label{eq:fine_objective}
\end{equation}
where $\mathcal{L}_{content}$ is $\ell_1$ loss used in image restoration.

\subsection{The derivation process and more details of the parameter regularization strategy}
Due to space limitations, we have simplified the derivation process of the parameter importance regularization strategy in the main paper. This article will introduce the derivation process and more details of the parameter importance regularization strategy in detail.

First, for simplicity, we denote the training for the diffuse generation task as Task 0 and the fine-tuning for the task-specific image restoration task as Task 1.
Assume that the training samples generated by diffusion and the training samples reconstructed by image restoration are denoted as $x_0$ and $x_1$ respectively. The network parameters trained on task 0 are denoted as $\theta^0=\theta_0^0,...,\theta_m^0$, while the network parameters trained on task 1 are denoted as $\theta^1=\theta_0^1,...,\theta_m^1$.
After training the network on Task 1, its performance degradation on the previous Task 0 can be evaluated by the following formula:
\begin{equation}
    \begin{split}        
\Delta f = f\left( {{x^0};{\theta ^1}} \right) - f\left( {{x^0};{\theta ^0}} \right) ,
    \end{split}
    \label{dataf}
\end{equation}
where $f$ denotes the network. Taking the element of parameter $\theta_k^0$ ($k$-th depth) for example, the change of parameter $\theta_k^0$ is denoted as $\delta {\theta_k^0}$ when model is trained on the new Task 1, the mathematical form is $\delta {\theta_k^0} = {\theta_k^1} - {\theta_k^0}$. Then, we take the Taylor expansion of $f\left( {{x^0};{\theta_k^1}} \right)$ at point $\theta_k^0$:
\begin{equation}
    \begin{split}        
f\left( {{x^0};\theta _k^1} \right) = f\left( {{x^0};\theta _k^0} \right) + {\left( {{\nabla _{\theta _k^0}}f} \right)^T} \cdot \theta _k^0 + \frac{1}{2}{\left( {\theta _k^0} \right)^T} \cdot \nabla _{\theta _k^0}^2f \cdot \theta _k^0 + O\left( {{{\left\| {\delta \theta _k^0} \right\|}^3}} \right).
    \end{split}
    \label{taylorf}
\end{equation}

Inspired by Gauss-Newton method, we approximate the $\nabla _{\theta _k^0}^2f$ to relieve the computational cost as:
\begin{equation}
    \begin{split}        
    \mathop {\mathbb E} \limits_{{x_0} \sim {\mathbb P^0}} \left[ {\nabla _{\theta _k^0}^2f} \right] \approx 2 \times \mathop {\mathbb E}\limits_{{x_0}\sim{\mathbb P^0}} {\left[ {\nabla f} \right]^T}\mathop {\mathbb E}\limits_{{x_0}\sim{\mathbb P^0}} \left[ {\nabla f} \right].
    \end{split}
    \label{expect_df}
\end{equation}

 And, we inject the Eq. \ref{taylorf} into Eq. \ref{dataf}, and acquire the weight importance as:
\begin{equation}
    \begin{split}        
 {I}{{\left( {\theta _k^0} \right)}} = \nabla f = {\left( {{\nabla _{\theta _k^0}}f} \right)^T} \cdot \delta \theta _k^0 + \frac{1}{2}{\left( {\delta \theta _k^0} \right)^T} \cdot {\left( {{\nabla _{\theta _k^0}}f} \right)^T} \cdot {\nabla _{\theta _k^0}}f \cdot \delta \theta _k^0 .
    \end{split}
    \label{eq:df}
\end{equation}
To maintain the performance of previous Task 0, we need to minimize the Eq. 1. From this motivation, when training the model on Task 1, we add a regularization term based on conventional loss to keep the knowledge of Task 0. In summary, the total loss on Task 1 is a composite loss, which is of the form:
\begin{equation}   
    \begin{split}  
\mathcal{L}'& = \mathcal{L} + \lambda \Delta f \\
& = \mathcal{L} + \frac{\lambda }{2}\sum\limits_{k = 1}^m {\left[ {{I}{{\left( {\theta _k^0} \right)}^T}\left| {\delta \theta _k^0} \right| + {{\left| {\delta \theta _k^0} \right|}^k}{{\left( {{\nabla _{\theta _k^0}}f} \right)}^T}\left| {\delta \theta _k^0} \right|} \right]} .
    \end{split}
    \label{eq:loss1}
\end{equation}
Note that the Eq. \ref{eq:loss1} is another form of Eq. \ref{importance} in the main body, which describes the implementation of the parameter regularization strategy more concretely.
The pseudo code of the paramter importance regularization strategy method is summarized in Algorithm \ref{alg:regularization}.

\begin{algorithm}[ht]
  \caption{Parameter Importance Regularization Strategy for Fine-tuning.} 
  \label{alg:regularization}
  \small
  \begin{algorithmic}[1]
    \Require Training diffusion generation phase task $T_0$, fine-tuning phase task $T_{1}$; conventional training loss $\mathcal{L}$
    \Require Final trained model parameters $\theta _k^1$ for Diffusion Training Enhanced IR Framework  

    \For{Task $T_0$}
      \State Training Task $T_0$ using conventional loss $\mathcal{L}$
          \If{last training epoch}
          \For{Parameter ${\theta _k^0}$}
          \State $\nabla f \leftarrow \overline {\left| {{\nabla _{\theta _k^0}}L} \right|} $
          \State $\theta _k^0 \leftarrow \theta _k^0 + \epsilon\nabla f $
          \EndFor  
          \EndIf
    \EndFor  
    \For{Task $T_1$}
      \State get conventional loss $\mathcal{L}_s$ through forward propagation
using $\mathcal{L}$
          \If{last training epoch}
          \For{Parameter ${\theta _k^1}$}
          \State $\delta \theta _k^0 \leftarrow \theta _k^1 - \theta _k^0  $
          \State ${I}{{\left( {\theta _k^0} \right)}} \leftarrow \nabla f $
          \State $\mathcal{L}_s \leftarrow \mathcal{L}_s + \frac{\lambda }{2}\sum\limits_{k = 1}^m {\left[ {{I}{{\left( {\theta _k^0} \right)}^T}\left| {\delta \theta _k^0} \right| + {{\left| {\delta \theta _k^0} \right|}^k}{{\left( {{\nabla _{\theta _k^0}}f} \right)}^T}\left| {\delta \theta _k^0} \right|} \right]} $
          \State $\theta _k^1 \leftarrow \theta _k^0 + \epsilon \nabla \mathcal{L}_s $
          \EndFor  
          \EndIf
    \EndFor  
    \State \textbf{return} $\theta _k^1$
  \end{algorithmic}
\end{algorithm}

\section{More Details}
\label{Details}
\subsection{Dataset and Experiment Setting}
\label{Dataset and Experiment Setting}
\begin{figure}[t]
    \centering
    \includegraphics[width=1.\linewidth]{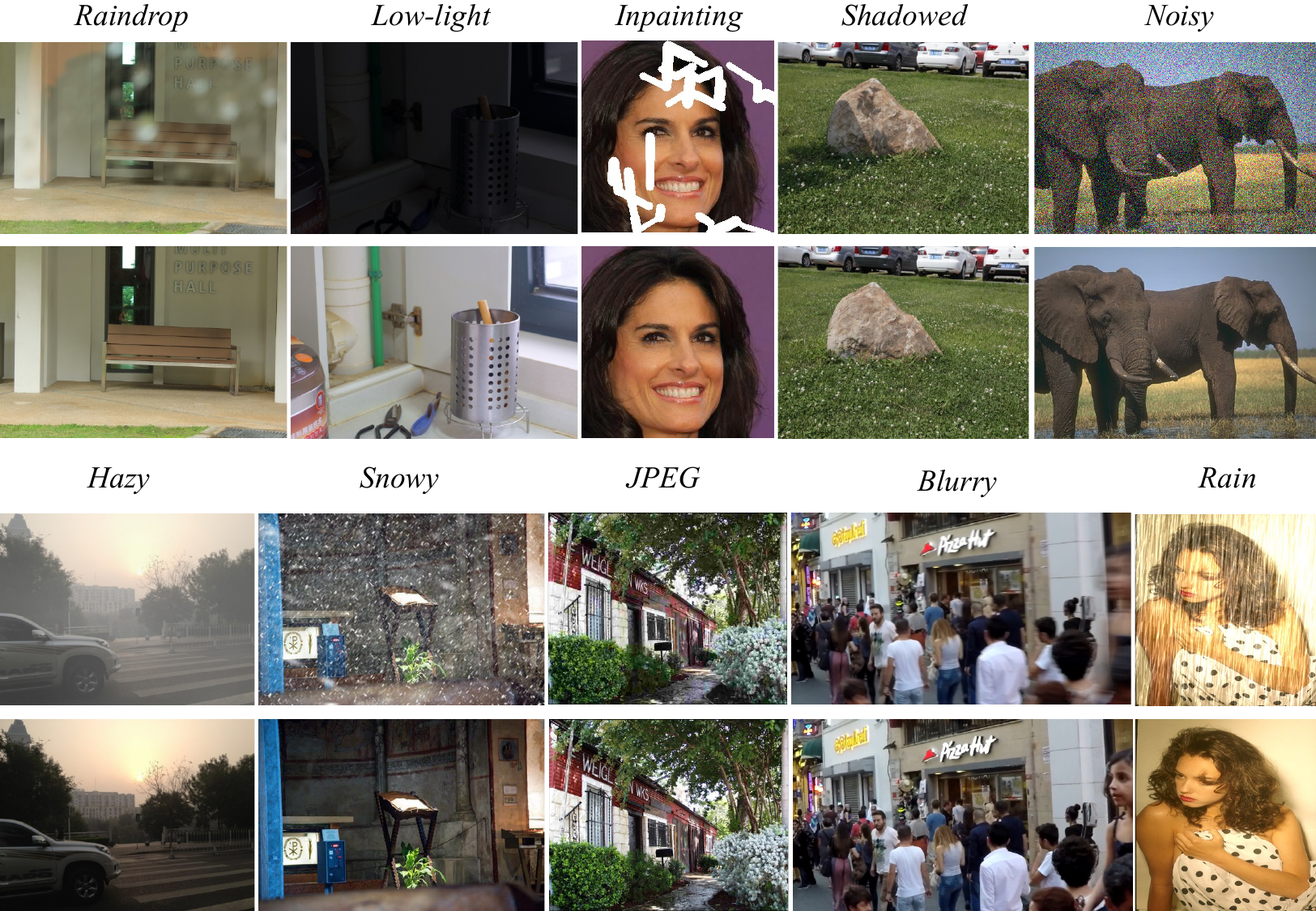}
    \caption{Example images with 10 image restoration tasks. For each task, the first row is the corrupted input and the second row is the result produced by our \emph{multi-task unified image restoration model}.}
    \label{fig:dataset-examples}
\end{figure}

In this section we elaborate on the mixed degradation dataset we use in multi-task unified image restoration (\textcolor{blue! 100}{Setting B}) .
The collected dataset contains ten different degradation types, including \emph{blurry}, \emph{hazy}, \emph{JPEG-compressing}, \emph{low-light}, \emph{noisy}, \emph{raindrop}, \emph{rainy}, \emph{shadowed}, \emph{snowy}, and \emph{inpainting}, as shown in \fref{fig:dataset-examples}. The details of these datasets are listed below:

\begin{itemize}
    \item \emph{Blurry}: collected from the GoPro~\citep{nah2017deep} dataset containing 2103 and 1111 training and testing images, respectively.
    \item \emph{Hazy}: collected from the RESIDE-6k~\citep{OTS2019} dataset which has mixed indoor and outdoor images with 6000 images for training and 1000 images for testing.
    \item \emph{JPEG-compressing}: the training dataset has 3440 images collected from DIV2K~\citep{agustsson2017ntire} and Flickr2K~\citep{timofte2017ntire}. The testing dataset contains 29 images from LIVE1~\citep{sheikh2005live}. Moreover, all LQ images are synthetic data with a JPEG quality factor of 10.
    \item \emph{Low-light}: collected from the LOL~\citep{wei2018deep} dataset containing 485 images for training and 15 images for testing.
    \item \emph{Noisy}: the training dataset is the same as that in \emph{JPEG-compressing} but all LQ images are generated by adding Gaussian noise with noise level 50. The testing images are from CBSD68~\citep{martin2001database} and also added that Gaussian noise.
    \item \emph{Raindrop}: collected from the RainDrop~\citep{qian2018attentive} dataset containing 861 images for training and 58 images for testing.
    \item \emph{Rainy}: collected from the Rain100H~\citep{yang2017deep} dataset containing 1800 images for training and 100 images for testing.
    \item \emph{Shadowed}: collection from the SRD~\citep{qu2017deshadownet} dataset containing 2680 images for training and 408 images for testing.
    \item \emph{Snowy}: collected from the Snow100K-L~\citep{Snow100K2018} dataset. Since the original dataset is too large (100K images), we only use a subset which contains 1872 images for training and 601 images for testing.
    \item \emph{Inpainting}: we use CelebaHQ as the training dataset and divide 100 images with 100 thin masks from RePaint~\citep{lugmayr2022repaint} for testing.
\end{itemize}

We also provide several visual examples for each task for a better understanding of the 10 degradations and datasets, as shown in \fref{fig:dataset-examples}.

Validation is conducted on 14 datasets using 2 experimental setups:

I.~~
{\textbf{Single-task IR and generalization}} (\textcolor{blue! 100}{Setting A})  

~~~~~~Dehazing from RESIDE~\cite{OTS2019} to REVIDE~\cite{9578789}, deraining from Rain100L~\cite{Rain100H_2017_CVPR} to Rain100H~\cite{Rain100H_2017_CVPR}, desnowing from Snow100K~\cite{Snow100K2018} to RealSnow~\cite{Zhu2023LearningWA}, low-light enhancement from LOL-v1~\cite{wei2018deep} to LOL-v2~\cite{Yang2021SparseGR}.

II.~~
{\textbf{Multi-task unified IR}} (\textcolor{blue! 100}{Setting B})

~~~~~~~10 restoration tasks as shown in \fref{fig:dataset-examples}. We use peak signal-to-noise ratio (PSNR) and structural similarity index (SSIM)~\citep{wang2004image} for fidelity evaluation, and the learned perceptual image patch similarity (LPIPS)~\citep{zhang2018unreasonable} for perceptual evaluation.

\subsection{Implementation Details }
\label{Implementation Details}
Our method is lightweight, making it feasible to train even with low-end graphics cards.
We implemented our method on the PyTorch platform using 8 NVIDIA GTX 1080Ti GPUs. The optimizer used is Adam with an exponential decay rate of 0.9. The initial learning rate is set to 5e-5, and a cosine annealing strategy is employed for adjustment. The patch size for images is set to 128 with the batch size of 16, and based on tuning experience, the hyperparameters are set as follows: $\alpha=0.2$, $a=0.05$, ${N_E=10}$. 

For the generative pre-training stage, we only perform generative training on clean images, and after 500K iterations, the restoration network acquires image generation capabilities. For fine-tuning of the image restoration task with enhanced generalization, we perform reconstruction training on paired images of different degradation types. This stage also undergoes 300K iterations for single-task image restoration, and 100K iterations for each incremental task of different degradation types for multi-task integrated image restoration. The final test will be conducted according to the settings of the benchmark test set.

\subsection{Detail of Timestep Matched Generative Pre-training}
\label{Detail of Timestep Matched Generative Pre-training}
We search for a matching time step $t$ during generative pre-training to match the degraded residual to the diffuse latent space. Specifically, we perform the reverse process during diffuse generative training to obtain the noise score predicted by the network, and then subtract the score of each time step from the degraded image, and find the time $t$ when the difference best matches the GT, which is the desired $t^{mat}$.
For the detailed process of obtaining the optimal matching time step during directional diffusion sampling, refer to Algorithm \ref{alg:sampling}.

\begin{table}[ht]
\centering
        \caption{Comparison of the number of parameters, model computational efficiency, and inference time. The flops and inference time are computed on face inpainting images of size $256 \times 256$. {Note that MAXIM is implemented with the JAX GPU version.}}
        \label{app-table:complexity}
        \begin{small}
        \begin{sc}
        \resizebox{.98\linewidth}{!}{
        \begin{tabular}{lcccccc}
        \toprule
    Merhod & MAXIM & RromptIR & NAFNet & Ours  & Ours + NAFNet & Ours + X-Restormer \\
    \#Param & 14.1M & 33M   & 67.8M & 27.8M & 67.8M + 1.8M & 89M + 1.8M \\
    Flops & 216G  & 158G  & 63G   & 72G   & 63G + 8G & 160G + 8G \\
    Runtime & 2.9s  & 0.1s  & 0.08s & 0.09s & 0.10s & 2.5s \\
        \bottomrule
        \end{tabular}}
        \end{sc}
        \end{small}
\end{table}

\section{Further Study Investigations}
\label{Further Study Investigations}

\subsection{Model Efficiency}
\label{Model Efficiency} 
We have shown the effectiveness of applying ours to various image restoration models and tasks. The test-time computational cost (FLOPs and runtime) is however virtually unaffected.
The quantitative results of the parameters and floating point numbers of the image restoration framework enhanced by diffusion training proposed by us and the parameters and floating point numbers of the framework combined with the existing image restoration network compared with the previous state-of-the-art methods are shown in Table \ref{app-table:complexity}.

\subsection{Additional Ablation Studies}
\label{Additional Ablation Studies} 
In the main text, we have given ablation studies for most of the components proposed in this paper, including generative pre-training with matching timesteps (TMGP), generalization-enhancing task-specific fine-tuning (GEF), MoE Adapters with Time-based Prompts (MoE), Time-sequential Incremental Training (TIT), parameter importance-guided parameter regularization strategy $L_{reg}$, and gradient orthogonalization techniques $L_{orthog}$ in the generation and reconstruction directions. Below, I will add a description of the matching time steps $t^{mat}$ and generation sensitivity-guided network-level regularization decay $w_{decay}(t)$. 

\subsubsection{{\textbf{Single-task IR and generalization}} (\textcolor{blue! 100}{Setting A})  }
For single-task image restoration and generalization tasks, we have conducted detailed ablation studies on all components using image dehazing as an example in the main text. In this section, we will complete all ablation studies on all remaining tasks: image deraining, image desnowing, and low-light image enhancement. In addition, we also provide additional explanations on the matching time $t^{mat}$ and network layer generation-sensitive regularization attenuation $w_{decay}(t)$, which are crucial to this study. Since MoE Adapters with Time-based Prompts (MoE) and Time-sequential Incremental Training (TIT) are only applied to multi-tasks, no additional analysis is required here.

The ablation study of the image dehazing task is shown in Table \ref{tab:Setting A dehazing}, the ablation study of the image deraining task is shown in Table \ref{tab:Setting A deraining}, the ablation study of the image desnowing task is shown in Table \ref{tab:Setting A desnowing}, and the ablation study of the low-light image enhancement task is shown in Table \ref{tab:Setting A Low-light}. From the vertical observation of the table, it can be found that all the proposed components have a promoting effect on the image restoration effect and the generalization effect. At the same time, the lack of a fine-tuning strategy for generalization enhancement significantly affects the generalization effect. 
Looking at the table horizontally, we can see that the various components in our proposed new framework have a strong promoting effect on the generalization of image restoration tasks.
Only when all components work together can the best generalization effect and image restoration effect be achieved. This confirms the effectiveness of the components proposed in this paper.

 \begin{table}[htbp]
  \centering
  \caption{Ablation study of image dehazing in {\textbf{Single-task IR and generalization}} (\textcolor{blue! 100}{Setting A}).}
    \label{tab:Setting A dehazing}%
  \setlength{\tabcolsep}{20pt}
    \begin{tabular}{c|cc|cc}
    \toprule [0.1em]
    \textcolor{blue! 100}{Image dehazing}  & \multicolumn{4}{c}{\textcolor{blue! 100}{Setting A}} \\
    \midrule
    \multirow{2}[4]{*}{\textbf{Method}} & \multicolumn{2}{c|}{Restoration} & \multicolumn{2}{c}{Generalization} \\
\cmidrule{2-5}          & PSNR↑ & SSIM↑ & PSNR↑ & SSIM↑ \\
    \midrule
    w/o TMGP & 28.95 & 0.895 & 15.29  & 0.535 \\
    w/o $t^{mat}$ & 28.63 & 0.844 & 15.04  & 0.513 \\
    w/o $w_{decay}(t)$ & 28.58 & 0.862 & 14.82  & 0.537 \\
    w/o GEF & 21.25 & 0.624 & 14.47  & 0.565 \\
    w/o $L_{reg}$ & 28.65 & 0.844 & 16.10  & 0.611 \\
    w/o $L_{orthog}$ & 28.79 & 0.909 & 19.88  & 0.742 \\
    \rowcolor{blue! 6} Ours  & \textbf{29.08} & \textbf{0.933} & \textbf{20.74 } & \textbf{0.782} \\
    \bottomrule [0.1em]
    \end{tabular}%
\end{table}%

 \begin{table}[htbp]
  \centering
  \caption{Ablation study of image deraining in {\textbf{Single-task IR and generalization}} (\textcolor{blue! 100}{Setting A}).}
    \label{tab:Setting A deraining}%
  \setlength{\tabcolsep}{20pt}
    \begin{tabular}{c|cc|cc}
    \toprule [0.1em]
    \textcolor{blue! 100}{Image deraining}  & \multicolumn{4}{c}{\textcolor{blue! 100}{Setting A}} \\
    \midrule
    \multirow{2}[4]{*}{\textbf{Method}} & \multicolumn{2}{c|}{Restoration} & \multicolumn{2}{c}{Generalization} \\
\cmidrule{2-5}          & PSNR↑ & SSIM↑ & PSNR↑ & SSIM↑ \\
    \midrule
    w/o TMGP & 36.66 & 0.949 & 13.96  & 0.563 \\
    w/o $t^{mat}$ & 36.51 & 0.947 & 13.84  & 0.585 \\
    w/o $w_{decay}(t)$ & 36.24 & 0.916 & 13.28  & 0.545 \\
    w/o GEF & 23.83 & 0.522 & 11.36  & 0.486 \\
    w/o $L_{reg}$ & 36.49 & 0.936 & 12.35  & 0.577 \\
    w/o $L_{orthog}$ & 36.52 & 0.954 & 15.62  & 0.651 \\
    \rowcolor{blue! 6}Ours  & \textbf{37.47} & \textbf{0.971} & \textbf{16.87 } & \textbf{0.695} \\
    \bottomrule [0.1em]
    \end{tabular}%
\end{table}%

\subsubsection{{\textbf{Multi-task unified IR}} (\textcolor{blue! 100}{Setting B})}
For the multi-task unified image restoration task, we conducted a detailed ablation study on the comprehensive performance of ten degradation types in the main text. In this section, we will further supplement the study on the importance of other components.
In addition, we also provide additional explanations on the matching time $t^{mat}$ and network layer generation-sensitive regularization attenuation $w_{decay}(t)$, which are crucial to this study.

Our ablation results for the multi-task unified image restoration task are shown in Table \ref{tab:Setting B} From the vertical observation table, we can see that the various components we proposed have a strong promoting effect on improving the multi-task integrated image restoration task, among which the task-specific fine-tuning with enhanced generalization plays a decisive role, and the hybrid expert adapter based on time step prompts can further enhance the sensory quality of the model. From the horizontal observation table, we can see that the proposed framework achieves the best results in both objective and sensory indicators. And the best results of objective quality and sensory quality are achieved when and only when all the proposed components work properly, which further proves the effectiveness of our proposed framework.

 \begin{table}[htbp]
  \centering
  \caption{Ablation study of image desnowing in {\textbf{Single-task IR and generalization}} (\textcolor{blue! 100}{Setting A}).}
    \label{tab:Setting A desnowing}%
  \setlength{\tabcolsep}{20pt}
    \begin{tabular}{c|cc|cc}
    \toprule [0.1em]
    \textcolor{blue! 100}{Image desnowing}  & \multicolumn{4}{c}{\textcolor{blue! 100}{Setting A}} \\
    \midrule
    \multirow{2}[4]{*}{\textbf{Method}} & \multicolumn{2}{c|}{Restoration} & \multicolumn{2}{c}{Generalization} \\
\cmidrule{2-5}          & PSNR↑ & SSIM↑ & PSNR↑ & SSIM↑ \\
    \midrule
    w/o TMGP & 33.13 & 0.931 & 24.34  & 0.507 \\
    w/o $t^{mat}$ & 33.25 & 0.923 & 24.54  & 0.526 \\
    w/o $w_{decay}(t)$ & 33.52 & 0.933 & 24.88  & 0.578 \\
    w/o GEF & 25.06 & 0.564 & 24.14  & 0.524 \\
    w/o $L_{reg}$ & 32.24 & 0.912 & 24.61  & 0.589 \\
    w/o $L_{orthog}$ & 32.16 & 0.905 & 26.64  & 0.789 \\
    \rowcolor{blue! 6} Ours  & \textbf{34.77} & \textbf{0.943} & \textbf{27.84 } & \textbf{0.829} \\
    \bottomrule [0.1em]
    \end{tabular}%
\end{table}%

 \begin{table}[htbp]
  \centering
  \caption{Ablation study of Low-light enhancement in {\textbf{Single-task IR and generalization}} (\textcolor{blue! 100}{Setting A}).}
    \label{tab:Setting A Low-light}%
  \setlength{\tabcolsep}{18pt}
    \begin{tabular}{c|cc|cc}
    \toprule [0.1em]
    \textcolor{blue! 100}{Low-light enhancement}  & \multicolumn{4}{c}{\textcolor{blue! 100}{Setting A}} \\
    \midrule
    \multirow{2}[4]{*}{\textbf{Method}} & \multicolumn{2}{c|}{Restoration} & \multicolumn{2}{c}{Generalization} \\
\cmidrule{2-5}          & PSNR↑ & SSIM↑ & PSNR↑ & SSIM↑ \\
    \midrule
    w/o TMGP & 22.98 & 0.811  & 21.04 & 0.636 \\
    w/o $t^{mat}$ & 21.56 & 0.736  & 22.35 & 0.781 \\
    w/o $w_{decay}(t)$ & 21.47 & 0.721  & 22.07 & 0.767 \\
    w/o GEF & 16.03 & 0.545  & 20.11 & 0.612 \\
    w/o $L_{reg}$ & 22.79 & 0.784  & 24.31 & 0.805  \\
    w/o $L_{orthog} $ & 22.84 & 0.798  & 24.52 & 0.827 \\
    \rowcolor{blue! 6} Ours  & \textbf{23.59} & \textbf{0.856} & \textbf{25.67} & \textbf{0.883} \\
    \bottomrule [0.1em]
    \end{tabular}%
\end{table}%

\begin{table}[htbp]
  \centering
  \caption{Ablation study of {\textbf{Multi-task unified IR}} (\textcolor{blue! 100}{Setting B})}
  \label{tab:Setting B}%
  \setlength{\tabcolsep}{26pt}
    \begin{tabular}{cccc}
    \toprule [0.1em]
   \multirow{3}[6]{*}{\textbf{Method}} & \multicolumn{3}{c}{\textcolor{blue! 100}{Setting B}} \\
\cmidrule{2-4}          & \multicolumn{3}{c}{Average performance} \\
\cmidrule{2-4}          & PSNR↑ & SSIM↑ & LPIPS↓ \\
      \midrule
  w/o $t^{mat}$ & 25.15 & 0.804 & 0.747 \\
    w/o $w_{decay}(t)$ & 25.58 & 0.783 & 0.549 \\
    w/o TMGP & 24.56 & 0.782 & 0.464 \\
    w/o TIT & 17.76 & 0.564 & 0.529 \\
    w/o MoE & 27.01 & 0.743 & 0.394 \\
    w/o GEF & 27.64 & 0.842 & 0.188 \\
    w/o $L_{reg}$ & 27.41 & 0.794 & 0.236 \\
    w/o $L_{orthog}$ & 27.25 & 0.812 & 0.252 \\
    \rowcolor{blue! 6} Ours  & \textbf{28.13} & \textbf{0.862} & \textbf{0.123} \\
    \bottomrule
    \end{tabular}%
  
\end{table}%

\subsection{Quantitative Evaluation }
\label{Quantitative Evaluation} 
 Similar to \cite{Zhu2023LearningWA,yang2023language}, we use peak signal-to-noise ratio (PSNR) and structural similarity index (SSIM)~\citep{wang2004image} for fidelity evaluation, and the learned perceptual image patch similarity (LPIPS)~\citep{zhang2018unreasonable} for perceptual evaluation.
PSNR and SSIM are calculated along the RGB channels.   
LPIPS is calculated using the pre-trained alexNet.

We presented visual performance of {\textbf{Single-task restoration and generalization}} (\textcolor{blue! 100}{Setting A})  with 4 weather conditions in Figure \ref{fig3}. Clearly, our results effectively retained background details while eliminating multiple weather factors.

In addition, we provide a visual comparison of the experimental results of \underline{\textbf{Multi-task unified IR}} (\textcolor{blue! 100}{Setting B}), as shown in \fref{fig4} and \fref{fig5}. It can be seen that compared with the previous state-of-the-art methods, our framework can more effectively remove different degradations in the image without changing and destroying the original Beijing structure of the image too much, which is crucial for image restoration models in real scenes.

\subsection{Additional Visualizations and Comparisons}
\label{Additional Visualizations and Comparisons}

In the following Figure \ref{fig3}-\ref{fig6}, we present additional visualization results. It can be observed that our proposed method excels in removing more degradation Artifacts compared to previous approaches, resulting in the restoration of clear background information.

\section{Future Work and Limitation}
\label{Future Work and Limitation.}
Although our method can be easily transferred to various general restoration networks and significantly improves the model's single-task image restoration generalization ability and multi-task unified image restoration effect, the diffusion-based training method still consumes a lot of resources in practical applications. Our future work will focus on lightweighting and enhancing this generative training strategy, designing a new generative pre-training paradigm to minimize the training time while retaining the model's generative ability, and further improving the single-task generalization ability and multi-task unified image restoration effect.

\section{Broader Impacts}
\label{Societal Impact.}
In today's world, image capture systems inevitably face a range of degradation issues, stemming from inherent noise in imaging devices, capture instabilities, to unpredictable weather conditions. Consequently, image restoration holds significant research and practical value. Our proposed Diffusion Training Enhanced IR Framework enables image restoration networks to refine degraded images more effectively. However, from a societal perspective, negative consequences may also arise. For instance, the actual texture discrepancies introduced by image restoration techniques could impact fair judgments in medical and criminal contexts. In such scenarios, it becomes imperative to integrate expert knowledge for making informed decisions.

\begin{figure}[htbp] 
    \centering
    \includegraphics[width=1\linewidth]{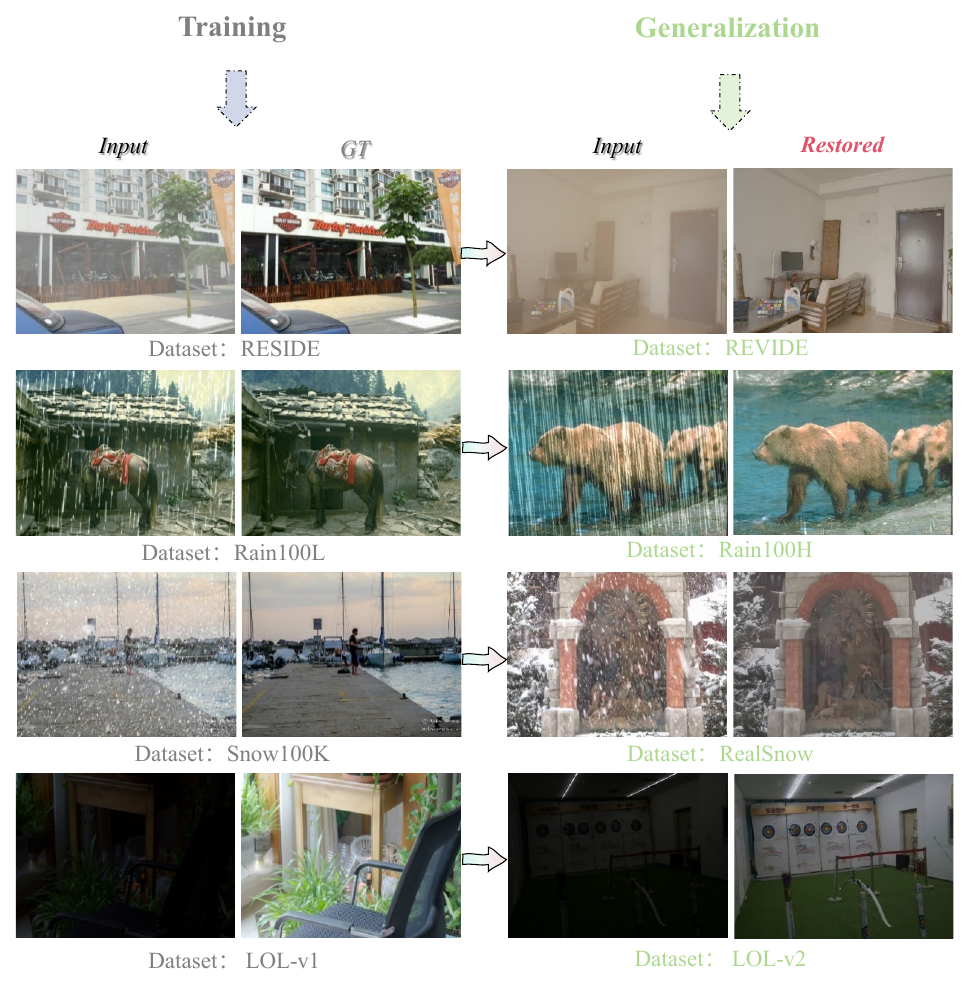}
    \caption{Visual performance of {\textbf{Single-task IR and generalization}} (\textcolor{blue! 100}{Setting A})  with 4 weather conditions. Our method can obtain cleaner images while preserving more background information when dealing with complex and multiple types of degradation.}
    \label{fig3}
\end{figure}

\begin{figure}[htbp] 
    \centering
    \includegraphics[width=1\linewidth]{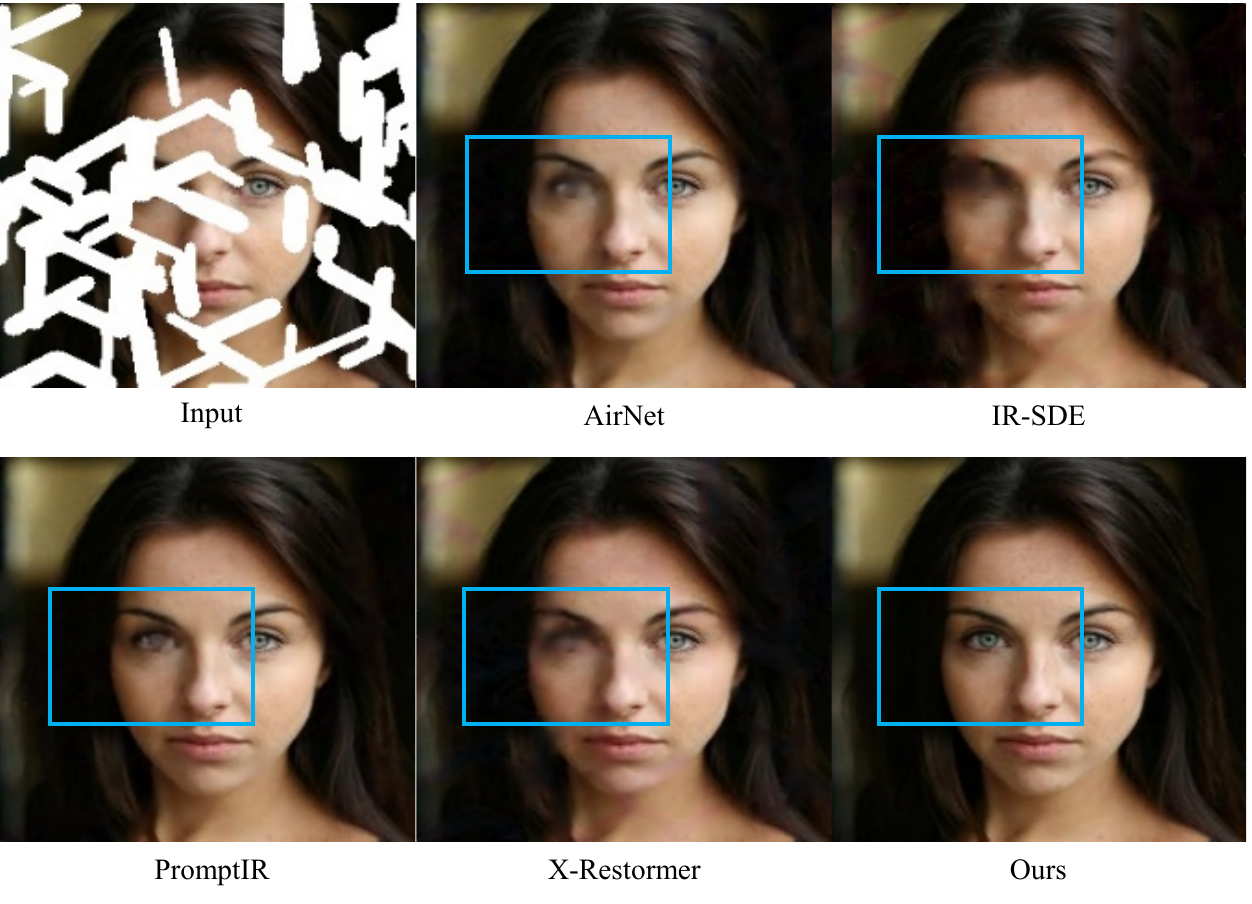}
    \caption{Image Inpainting visualization comparisons of our method with previous approaches for {\textbf{Multi-task unified IR}} (\textcolor{blue! 100}{Setting B}). Our method can obtain cleaner images while preserving more background information when dealing with complex and multiple types of degradation.}
    \label{inpaint_1}
\end{figure}

\begin{figure}[htbp] 
    \centering
    \includegraphics[width=1\linewidth]{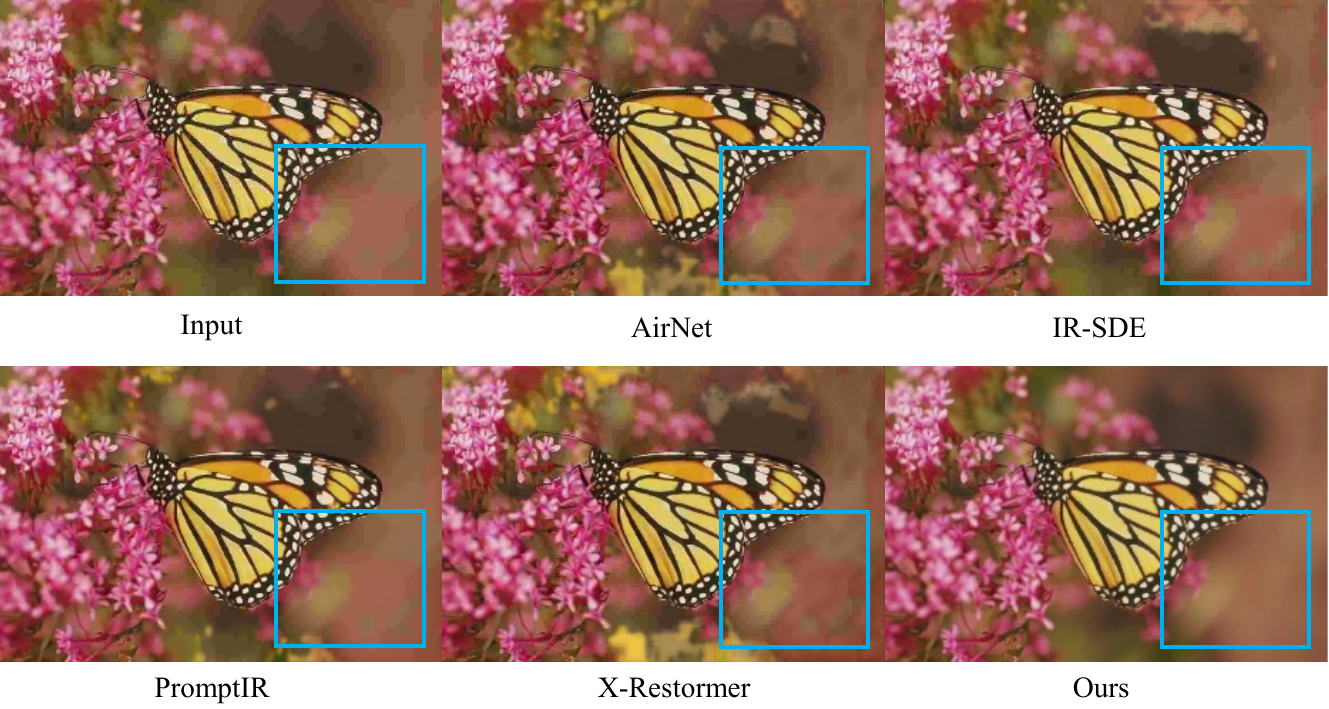}
    \caption{JPEG artifact deduction visualization comparisons of our method with previous approaches for {\textbf{Multi-task unified IR}} (\textcolor{blue! 100}{Setting B}). Our method can obtain cleaner images while preserving more background information when dealing with complex and multiple types of degradation.}
    \label{jpeg}
\end{figure}

\begin{figure}[htbp] 
    \centering
    \includegraphics[width=1\linewidth]{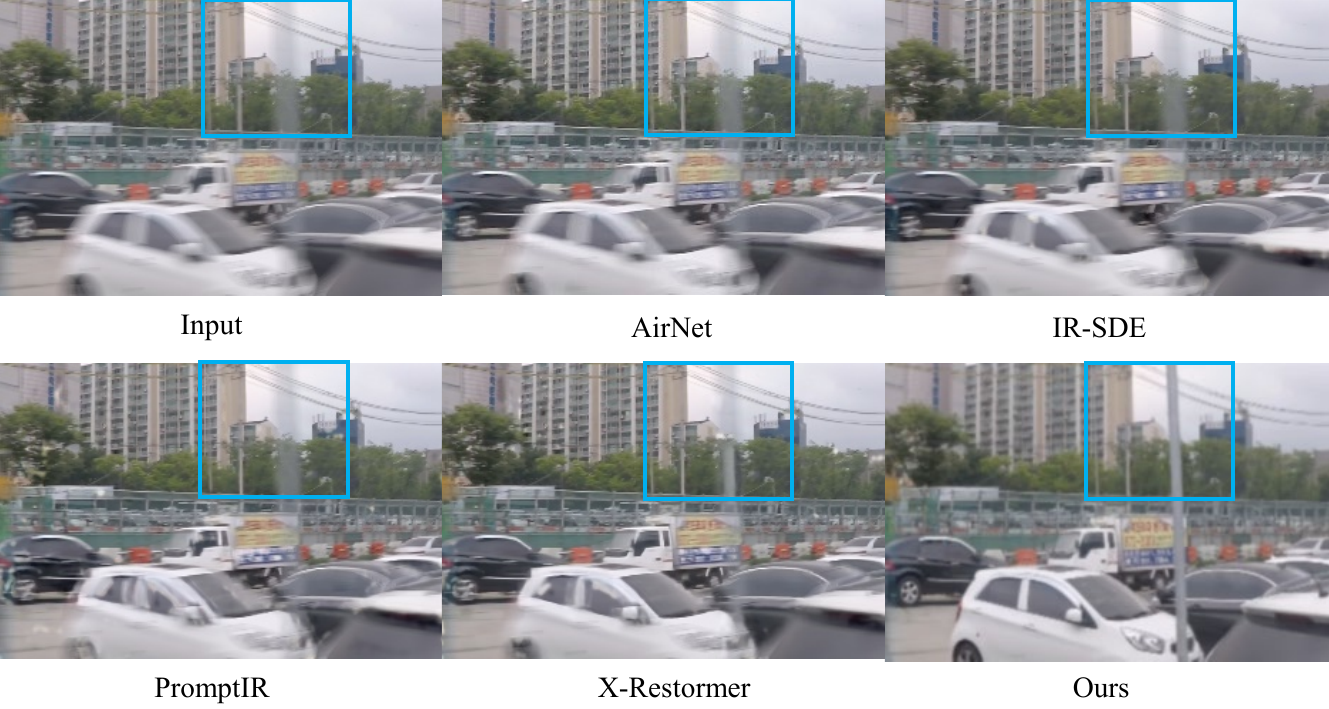}
    \caption{Image deblurring visualization comparisons of our method with previous approaches for {\textbf{Multi-task unified IR}} (\textcolor{blue! 100}{Setting B}). Our method can obtain cleaner images while preserving more background information when dealing with complex and multiple types of degradation.}
    \label{blur}
\end{figure}

\begin{figure}[htbp] 
    \centering
    \includegraphics[width=1\linewidth]{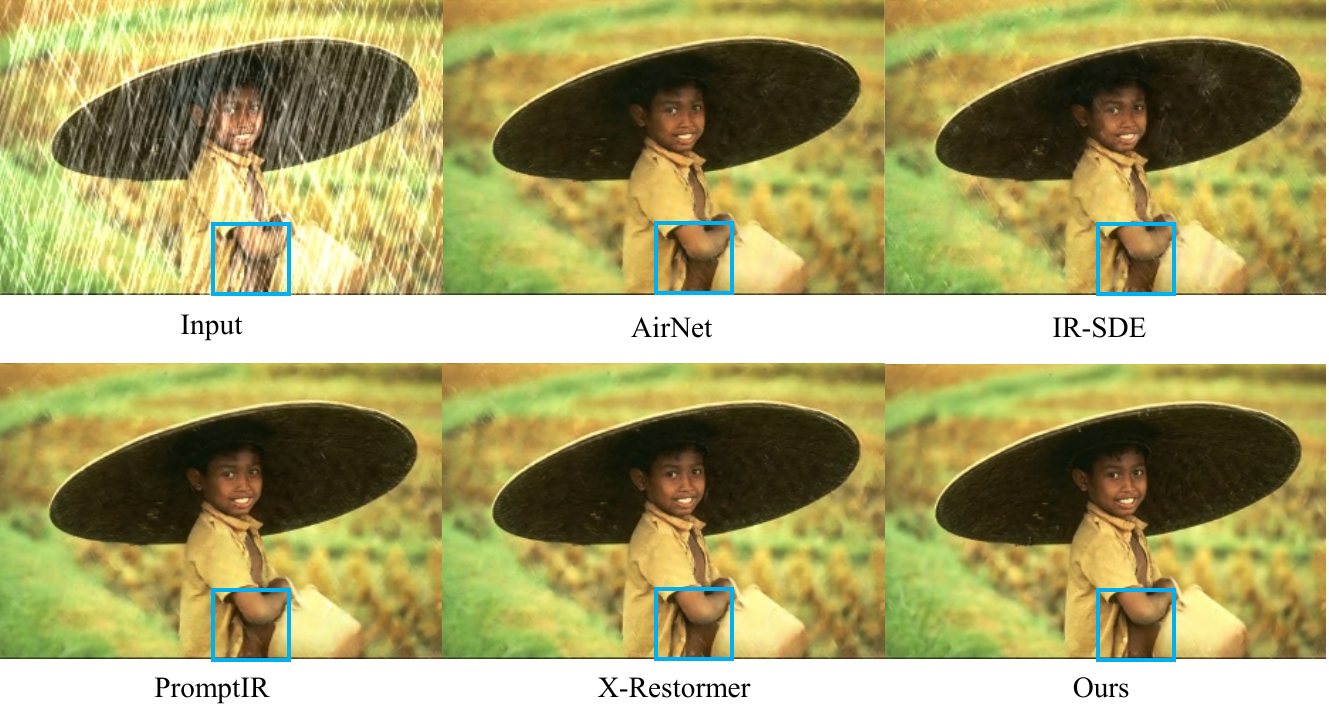}
    \caption{Image deraining visualization comparisons of our method with previous approaches for {\textbf{Multi-task unified IR}} (\textcolor{blue! 100}{Setting B}). Our method can obtain cleaner images while preserving more background information when dealing with complex and multiple types of degradation.}
    \label{rain}
\end{figure}

\begin{figure}[htbp] 
    \centering
    \includegraphics[width=1\linewidth]{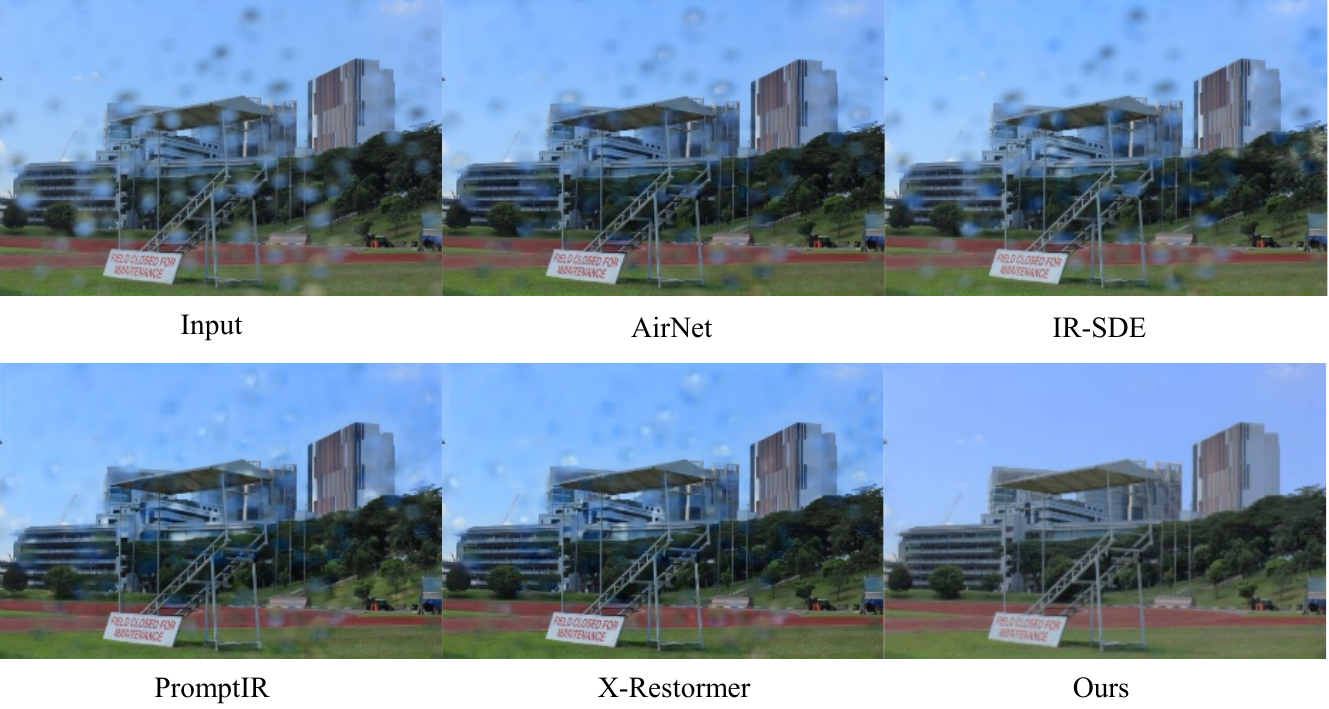}
    \caption{Image raindrop removal visualization comparisons of our method with previous approaches for {\textbf{Multi-task unified IR}} (\textcolor{blue! 100}{Setting B}). Our method can obtain cleaner images while preserving more background information when dealing with complex and multiple types of degradation.}
    \label{raindrop}
\end{figure}

\begin{figure}[htbp] 
    \centering
    \includegraphics[width=1\linewidth]{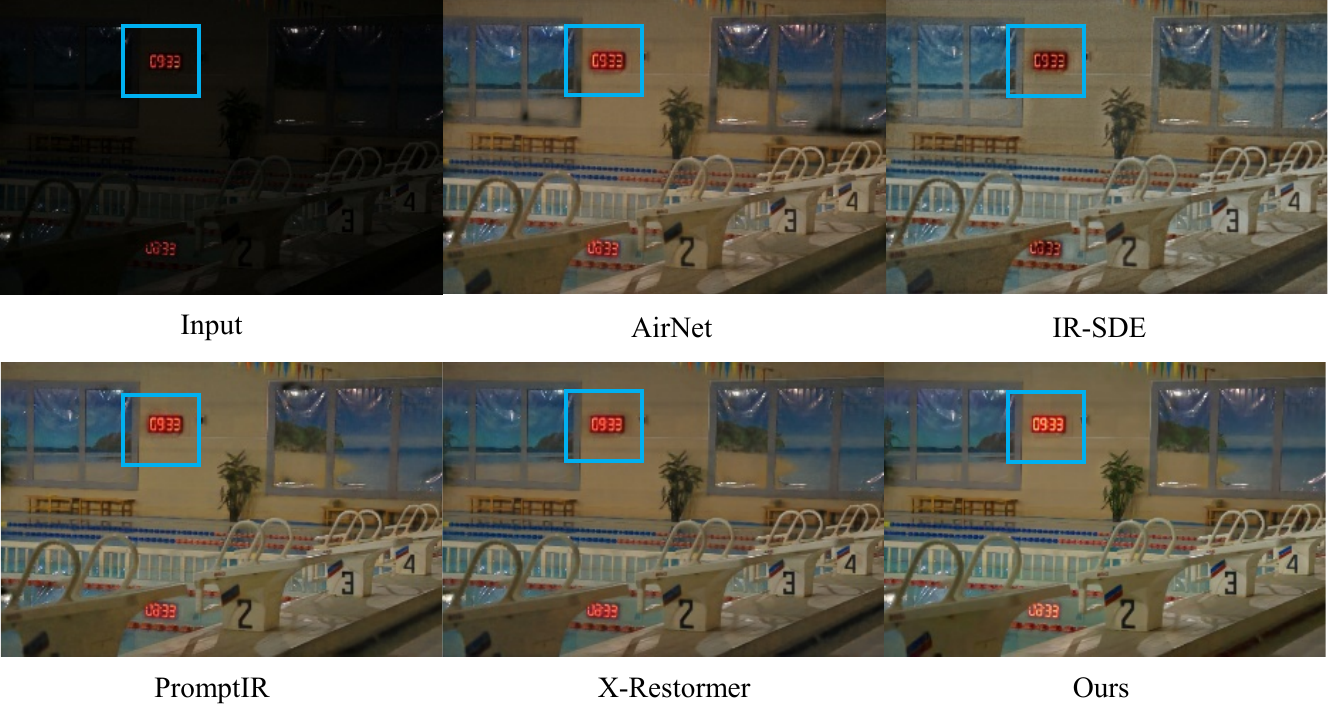}
    \caption{Low-light image enhancement visualization comparisons of our method with previous approaches for {\textbf{Multi-task unified IR}} (\textcolor{blue! 100}{Setting B}). Our method can obtain cleaner images while preserving more background information when dealing with complex and multiple types of degradation.}
    \label{Low-light}
\end{figure}

\begin{figure}[htbp] 
    \centering
    \includegraphics[width=1\linewidth]{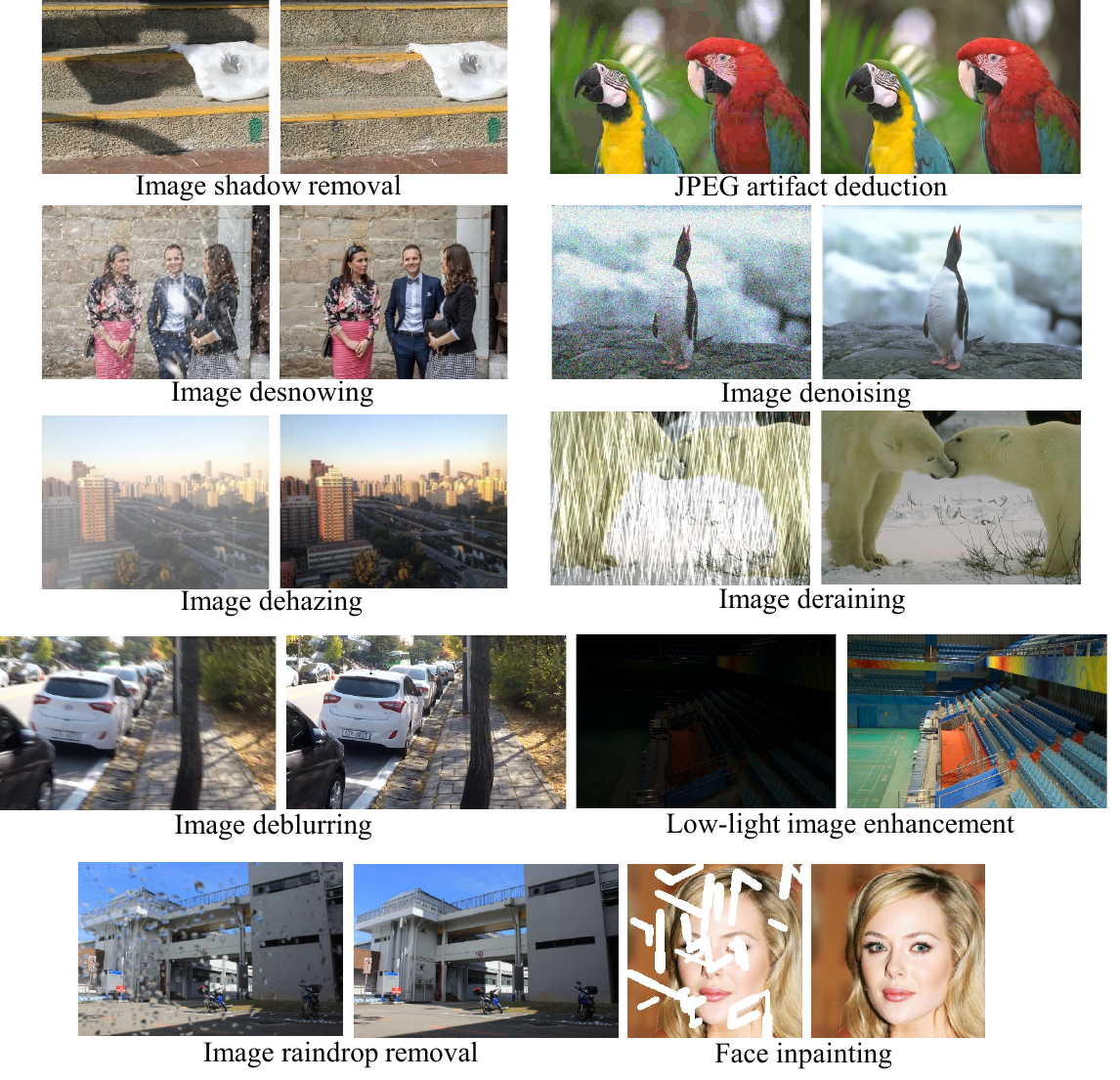}
    \caption{Visualization of our method for {\textbf{Multi-task unified IR}} (\textcolor{blue! 100}{Setting B}). Our method can obtain cleaner images while preserving more background information when dealing with complex and multiple types of degradation.}
    \label{muti}
\end{figure}

\begin{figure}[htbp] 
    \centering
    \includegraphics[width=1\linewidth]{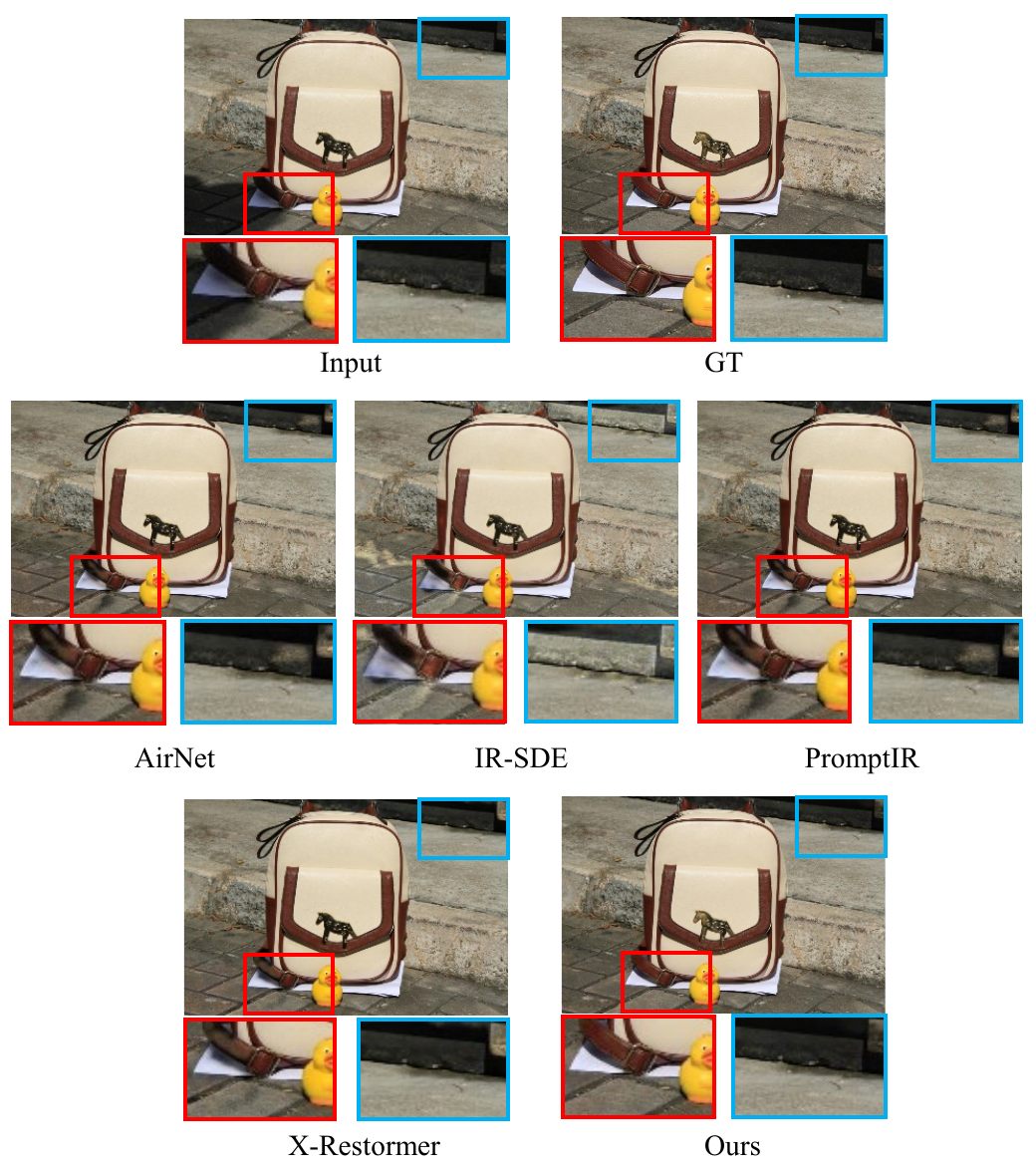}
    \caption{Visualization comparisons of our method with previous approaches for {\textbf{Multi-task unified IR}} (\textcolor{blue! 100}{Setting B}). Our method can obtain cleaner images while preserving more background information when dealing with complex and multiple types of degradation.}
    \label{fig4}
\end{figure}

\begin{figure}[htbp] 
    \centering
    \includegraphics[width=1\linewidth]{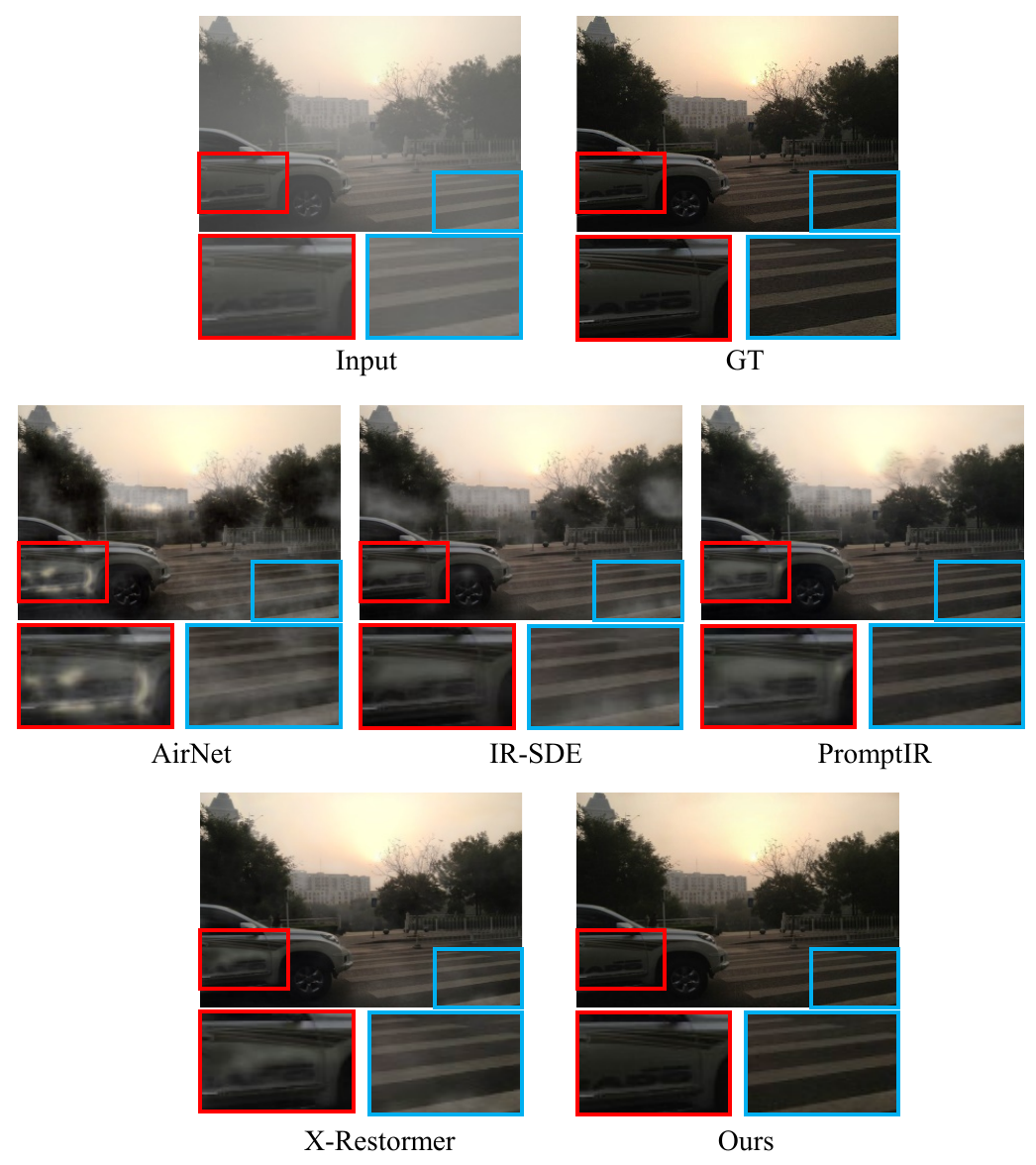}
    \caption{Visualization comparisons of our method with previous approaches for {\textbf{Multi-task unified IR}} (\textcolor{blue! 100}{Setting B}). Our method can obtain cleaner images while preserving more background information when dealing with complex and multiple types of degradation.}
    \label{fig5}
\end{figure}

\begin{figure}[htbp] 
    \centering
    \includegraphics[width=1\linewidth]{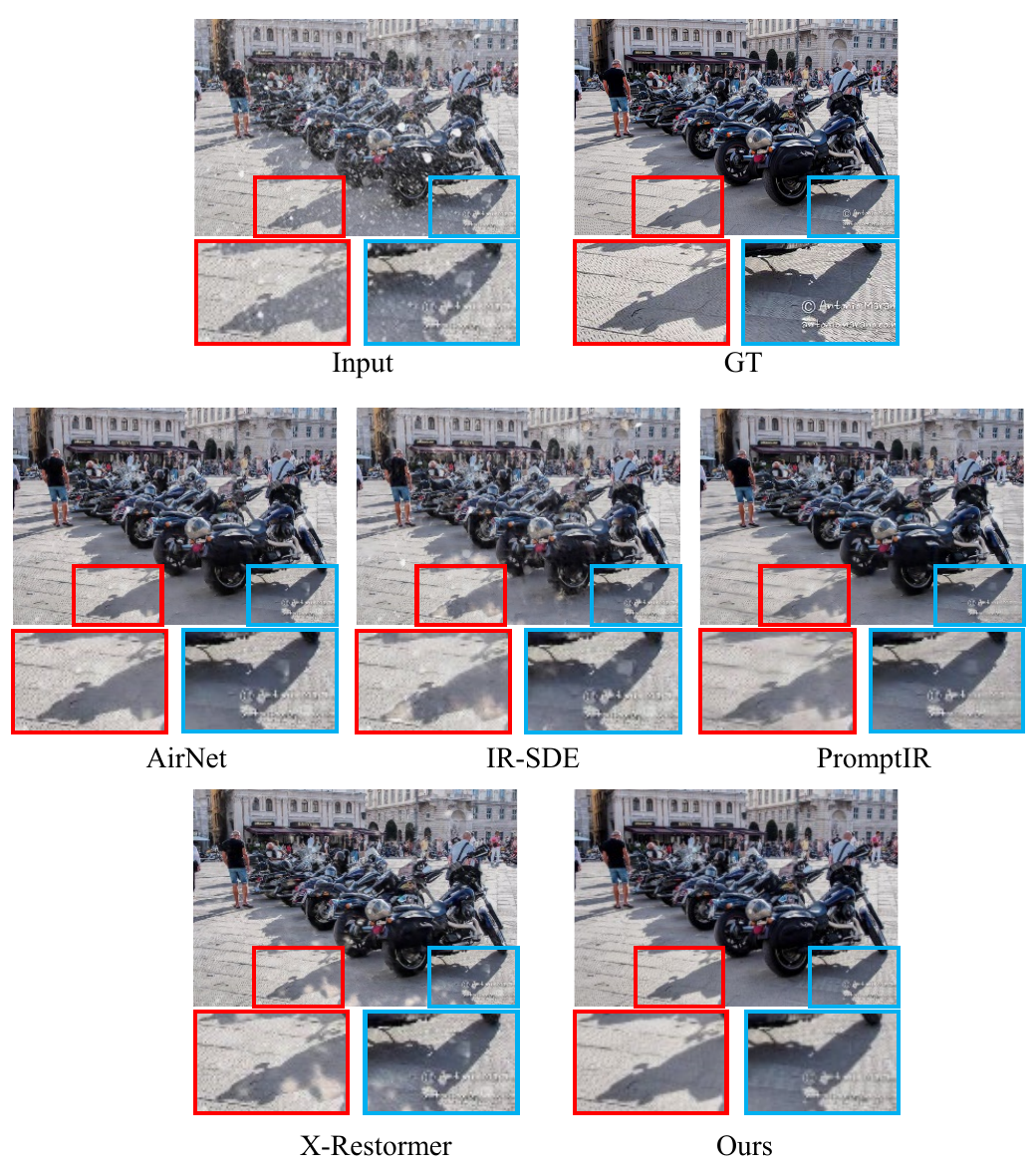}
    \caption{Visualization comparisons of our method with previous approaches for {\textbf{Multi-task unified IR}} (\textcolor{blue! 100}{Setting B}). Our method can obtain cleaner images while preserving more background information when dealing with complex and multiple types of degradation.}
    \label{fig6}
\end{figure}

\clearpage
{
\small
\bibliographystyle{abbrvnat}
\bibliography{main}
}

\end{document}